\documentclass{article}

     \PassOptionsToPackage{numbers, sort}{natbib}
    \usepackage[preprint]{neurips_2020}

\usepackage[utf8]{inputenc} % allow utf-8 input
\usepackage[T1]{fontenc}    % use 8-bit T1 fonts
\usepackage{hyperref}       % hyperlinks
\usepackage{url}            % simple URL typesetting
\usepackage{booktabs}       % professional-quality tables
\usepackage{amsfonts}       % blackboard math symbols
\usepackage{nicefrac}       % compact symbols for 1/2, etc.
\usepackage{microtype}      % microtypography
\usepackage{color}
\usepackage{graphicx}
\usepackage{epsfig}
\usepackage{subfigure}
\usepackage{amsmath,amsfonts,bm}

\def\eqref#1{equation~\ref{#1}}
\def\1{\bm{1}}

\def\rz{{\textnormal{z}}}

\def\rvu{{\mathbf{i}}}

\def\rvu{{\mathbf{u}}}

\def\rvx{{\mathbf{x}}}

\def\rvz{{\mathbf{z}}}

\def\vx{{\bm{x}}}

\def\vz{{\bm{z}}}

\def\mI{{\bm{I}}}

\DeclareMathAlphabet{\mathsfit}{\encodingdefault}{\sfdefault}{m}{sl}
\SetMathAlphabet{\mathsfit}{bold}{\encodingdefault}{\sfdefault}{bx}{n}

\newcommand{\E}{\mathbb{E}}

\newcommand{\R}{\mathbb{R}}

\newcommand{\KL}{D_{\mathrm{KL}}}

\newtheorem{theorem}{Theorem}
\newtheorem{lemma}{Lemma}

\newtheorem{assumption}{Assumption}

\title{Towards Better Understanding of Disentangled Representations via Mutual Information}
\author{%
	Xiaojiang Yang\\
	Shanghai Jiao Tong University\\
	\texttt{yangxiaojiang@sjtu.edu.cn} \\
	\And
	Wendong Bi\\
	Shanghai Jiao Tong University\\
	\texttt{biwendong@sjtu.edu.cn} \\
	\And
	Yitong Sun \\
	Huawei Noah's Ark Lab \\
	\texttt{sunyitong@huawei.com} \\
	\AND
	Yu Cheng\\
	Microsoft\\
	\texttt{yu.cheng@microsoft.com} \\
	\And
	Junchi Yan\thanks{Corresponding author} \\
	Shanghai Jiao Tong University\\
	\texttt{yanjunchi@sjtu.edu.cn} \\
}

\begin{document}

\maketitle
%Towards a more comprehensive description of disentanglement
\begin{abstract}
  Most existing works on disentangled representation learning are solely built upon an marginal independence assumption: all factors in disentangled representations should be statistically independent. This assumption is necessary but definitely not sufficient for the disentangled representations without additional inductive biases in the modeling process, which is shown theoretically in recent studies. We argue in this work that disentangled representations should be characterized by their relation with observable data. In particular, we formulate such a relation through the concept of mutual information: the mutual information between each factor of the disentangled representations and data should be invariant conditioned on values of the other factors. Together with the widely accepted independence assumption, we further bridge it with the conditional independence of factors in representations conditioned on data. Moreover, we note that conditional independence of latent variables has been imposed on most VAE-type models and InfoGAN due to the artificial choice of factorized approximate posterior $q(\rvz|\rvx)$ in the encoders. Such an arrangement of encoders introduces a crucial inductive bias for disentangled representations. To demonstrate the importance of our proposed assumption and the related inductive bias, we show in experiments that violating the assumption leads to decline of disentanglement among factors in the learned representations.
\end{abstract}

\section{Introduction}
Learning disentangled representations has been considered as an important step towards interpretable and more efficient machine learning~\cite{bengio2013representation, bengio2007scaling, lake2017building, schmidhuber1992learning, tschannen2018recent}. The disentangled representations are demonstrated to be interpretable or semantically meaningful~\cite{chen2016infogan,kumar2018variational}, robust to adversarial attacks~\cite{alemi2017deep} and related to fairness ~\cite{locatello2019on}. They are also useful for many downstream tasks, including sequential data generating~\cite{yingzhen2018disentangled}, reinforcement learning~\cite{higgins2017darla,nair2018visual}, robot learning~\cite{adrien2018curiosity}, transfer~\cite{liu2018a} and few shot learning~\cite{janzing2012on,bengio2013representation}, etc.

There are many attempts for learning disentangled representations, most of which are based on generative adversarial nets (GANs)~\cite{goodfellow2014generative} and variational auto-encoders (VAEs)~\cite{kingma2014auto-encoding}. InfoGAN~\cite{chen2016infogan} aims at disentangling the factors of variation in images by recovering factorized distributed latent variables from the generated images. VAE-based models for learning disentangled representations have been proposed from different motivations, such as limiting the bottleneck capacity~\cite{higgins2017beta-vae,burgess2018understanding}, penalizing the total correlation (see Eq.~\ref{eq:tc})~\cite{kim2018disentangling,chen2018isolating}, and matching the marginal with factorized prior~\cite{kumar2018variational}. 

These models all try to learn representations with factorized distributions~\cite{locatello2019Challenging,locatello2018competitive,kim2018disentangling,chen2018isolating}, in order to obtain disentangled representations. However, it is impossible to learn disentangled representations in unsupervised manner solely based on assumptions posed on marginal distributions of representations without additional inductive biases on models and data distributions~\cite{hyvarinen1999nonlinear,locatello2019on,khemakhem2019variational}. 
So, what are the missing inductive biases in our modeling process?
We note that the well-known description of disentangled representations given by Bengio et al.~\cite{bengio2013representation} is never just about the distribution of representations. Instead, the emphasized property is stated on the interaction between factors of disentangled representations and the data: a representation is disentangled if each factor corresponds to a single factor of variation in data, and meanwhile is invariant to other factors of variation~\cite{kim2018disentangling,locatello2019Challenging,higgins2017beta-vae}. 
However, how to describe the interaction between the factors and the data is hard since we usually do not have knowledge about the distribution of the data. To overcome this difficulty, we utilize mutual information between factors of representations and data to characterize disentangled representations.
 
% We argue that more assumptions are necessary and substantial for disentanglement due to two considerations: (1) As in Bengio's statement~\cite{bengio2013representation}, disentanglement is the bijective mapping between factors in representations and factors of variation in data, which emphasizes the relation between data and representations, and hence the common assumption is insufficient for disentanglement as it is merely about the intrinsic property of disentangled representations; (2) New assumptions can induce new constraints on models and data sets, which lead to inductive biases for disentanglement and might explain the disentanglement in previous works.

In particular, in addition to the assumption of independently distributed factors of representations, we propose another assumption for disentanglement: 

\emph{the mutual information between data and each factor in disentangled representations is invariant with respect to other factors.} 

Our assumption is a natural implication of Bengio's statement on disentanglement. For example, consider the construction of the dataset of dSprites~\cite{higgins2017beta-vae} and regard the underlying factors as a disentangled representation of the data set. For any given shape, x position, y position and rotation angle, within a certain range, each value taken by the scale factor always corresponds to a unique object. Intuitively, different underlying factors retain distinct information about data.
%This assumption is intuitively necessary but not sufficient for Bengio's statement, specifically not sufficient to ensure the bijective mapping between factors in representations and factors of variation in data. Nevertheless, the proposed assumption does bring further understanding for disentanglement and lead to an inductive bias for disentanglement as will be shown in the paper. Note that in this work, we focus on the case that the latent variables are stochastic when any data point is given, because only in this case, the mutual information is well-defined.

To understand how the proposed assumption (Eq.~\ref{eq:assumption2}) affects our modeling process, we show that it implies a factorized form of the distribution of the representation conditioned on data when the marginal independence assumption (Eq.~\ref{eq:assumption1}) is also posed. In other words, the factors of disentangled representations are independent conditioned on data. % $x$.
%Then we theoretically bridge it with independence and conditional independence of factors in representations, showing that conditional independence is also a necessary property for disentanglement. Motivated by this, we show that 
Interestingly, such a conditional independence requirement in the modeling implied by our assumption is actually satisfied in many representation learning models. For example, following Kingma's seminal work~\cite{kingma2014auto-encoding}, most VAE type representation learning algorithms choose factorized noise in the reparameterization. We believe that this actually introduces an important inductive bias for disentanglement. This result partly explains why disentanglement arises in models with such choice, including InfoGAN~\cite{chen2016infogan} and many VAEs~\cite{kingma2014auto-encoding,higgins2017beta-vae,kim2018disentangling,chen2018isolating,zhao2019infovae,kumar2018variational}. 

We show in experiments that violating the conditional independence by involving correlated approximate posterior leads to decline of disentanglement. For vanilla VAE~\cite{kingma2014auto-encoding}, FactorVAE~\cite{kim2018disentangling} and TCVAE~\cite{chen2018isolating}, correlated approximate posteriors lead to highly entangled representations for data sets of dSprites~\cite{higgins2017beta-vae}, SmallNORB~\cite{lecun2004learning} and Cars3D~\cite{reed2015deep}. This is quantified using Mutual Information Gap (MIG)~\cite{chen2018isolating}. Similar effects are observed for InfoGAN~\cite{chen2016infogan} by visualizing generated images of MNIST~\cite{lecun1998gradient} with factors of representations traversed. These results validate the necessity of our proposed assumption for disentangled representations.

Our main contributions can be summarized as follows:
\begin{itemize}
    \item We propose a fundamental assumption for disentanglement via mutual information in addition to the marginal independence. And we bridge it with the conditional independence among factors of representations conditioned on data. 
    % \item Based on our theoretical analysis, we show that factorized noise in parameterization ensure conditional independence, which is an inductive bias for disentanglement.
    \item We conjecture that the partial success of efforts in learning disentangled representations stems from the introduction of an inductive bias: the factorized approximate posterior. Our experimental results on VAEs and InfoGAN show that violating the conditional independence by using correlated approximate posteriors leads to decline of disentanglement, and thus support our conjecture and the importance of our proposed assumption.
\end{itemize}
\textbf{Notation.}
Throughout the paper, we denote data by the random vector $\rvx$, and their representations by another random vector $\rvz\in \R^J$. Each factor in the representation $\rvz$ is denoted by $z_j$ for $j\in[J]$, where $[J]=\{1, 2, \dots, J\}$. The joint distribution of encoders is denoted as $q(\rvx,\rvz)$, in which $q(\rvz|\rvx)$ is referred as the approximate posterior, and $q(\rvz)$ is the marginal.

All the proofs can be found in the supplementary material.

\section{Related Works}\label{sec:related}
There are many works about disentanglement through various approaches. A line of works~\cite{yingzhen2018disentangled,denton2017unsupervised,hsu2017unsupervised} focus on disentangled representations from common data sets like images, videos etc. Attention is also paid to supervised learning~\cite{mathieu2016disentangling} or semi-supervised learning of disentangled representations~\cite{spurr2017guiding,siddharth2017learning}. And others explore some novel loss functions for learning disentangled representations~\cite{rubenstein2018learning,zhao2019infovae,ridgeway2018learning}. There are also works on evaluation of disentanglement~\cite{eastwood2018a,do2020theory}. Since our assumption and analysis is for the learnability of disentangled representation learning, in next we will focus on works related to unsupervised learning of disentangled representations, definition of disentanglement, and identifiability of nonlinear ICA.

Unsupervised learning of disentangled representations has been proposed in early works on generative models. The authors in~\cite{schmidhuber1992learning} proposed a variant of auto-encoder to learn disentangled representations by minimizing the predictability of one factor in representations when other factors are fixed. This model is motivated by the independence among factors in the representation.
%, i.e. the common assumption. 
\cite{desjardins2012disentangling} and~\cite{reed2014learning} proposed variants of (Restricted) Boltzmann Machine in which interactions act to entangle the factors. 

Recent studies on unsupervised learning of disentangled representations are mainly based on GANs and VAEs. In the line of GANs, InfoGAN~\cite{chen2016infogan} penalizes the mutual information of representations, and qualitatively shows that different factors in representations correspond to different visual concepts. The authors in~\cite{brakel2018learning} propose to penalize the Jensen-Shannon divergence between the distribution of representations and its factorized distribution with a discriminator, based on Independent Component Analysis (ICA).

The most popular unsupervised models for disentangled representation learning are variants of vanilla VAE due to its stability. $\beta$-VAE~\cite{higgins2017beta-vae} encourages the encoder to learn disentangled representations by penalizing the KL term in the objective of vanilla VAE. Annealed VAE~\cite{burgess2018understanding} proposes to progressively increase the bottleneck capacity of VAE to encourage the encoder to learn different factors of variation when capacity grows. FactorVAE~\cite{kim2018disentangling} uses a discriminator to penalize the total correlation via ratio trick to enhance independence of factors in representations. DIP-VAE~\cite{kumar2018variational} matches the distribution of representations with disentangled priors. In TCVAE~\cite{chen2018isolating}, the authors decompose the objective of VAE and argue that the total correlation term is the source of disentanglement, then they derive a mini-batch estimator for the total correlation term and penalize it to enhance disentanglement. Most VAE-based models can be attributed to penalizing the total correlation and thus enhancing the independence among factors of representations.
%, which coincides with the common assumption.

Recently, there are several works~\cite{ridgeway2018learning,higgins2018towards,eastwood2018a,shu2020weakly,do2020theory} aiming at defining disentanglement, among which \cite{do2020theory} is most relevant to our work. The authors define disentangled representations through two properties: (i) The mutual information between each pair of factors in a disentangled representation should be zero, which is equivalent to the independence among factors of representations. And (ii) for each factor in a disentangled representation, there exists a factor of variation such that their mutual information is equal to their entropy, which is a formulation of Bengio's statement~\cite{bengio2013representation}. The second property describes the relation between factors in representations with factors of variation. In unsupervised scenario, however, the ground truth of factors of variation is not available, and hence this definition cannot lead to any practical guidance for disentangled representation learning. The authors also mention a relation between data and disentangled representations via multivariate mutual information~\cite{mcgill1954multivariate}, but no further analysis and inductive biases are discussed.

Another important line of works connect disentanglement to identifiability problems of non-linear ICA. For an identifiable model, it is possible to learn from data the representation that corresponds to the prior factors in the underlying generative models~\cite{hyvarinen1999nonlinear}. \citet{khemakhem2019variational} prove an important non-linear identifiability theorem, by assuming factorized priors conditioned on an extra observable variable $\rvu$. Under such an assumption, Flow-based models ~\cite{sorrenson2020disentanglement} are proposed to learn disentangled representations and achieve impressive experimental results. Different from these works, we provide necessary instead of sufficient conditions for disentangled representations, and we do not need to assume the unique underlying generative model for data. However, we conjecture that there are deeper connections between our assumption and the identifiability results. If we treat the extra observable variable $\rvu$ in identifiability results as a part of data, their assumption of the factorized form of $q(\rvz|\rvu)$ coincides with the conditional independence of $\rvz$ on $\rvx$ derived from our proposed assumptions. 
% These works demonstrate the importance of factors' independence in representations conditioned on an additionally observed variable for disentanglement, which is similar with our conclusion.

%Impossibility theorem~\cite{locatello2019Challenging} is also an important result for unsupervised learning of disentangled representations. It claims that without inductive biases on both models and data sets, it is impossible to learn disentangled representations with unsupervised manner by simply ensuring independence of factors. Note that the proposed assumption does not simply focus on the independence of factors. Whether it breaks the impossibility theorem need be further investigated as future work.

% \section{Theoretical Analysis}
\section{Proposed Assumption for Disentanglement}
In this section, we first review the marginal independence assumption for disentanglement, pointing out that it merely describes the intrinsic property for representations and thus not enough for learning disentangled representations. To describe the relation between data and representations, we propose another assumption via mutual information.

\subsection{Existing Independence Assumption}
As summarized in Section~\ref{sec:related}, most recent unsupervised models for disentangled representation learning penalize the total correlation of marginal distribution:
\begin{equation}
TC(\rvz)\equiv\KL\left(q(\rvz)\Vert \prod_{j=1}^J q(\rz_i)\right) \label{eq:tc}
\end{equation}
where $q(\rvz)=\E_{q(\rvx)}[q(\rvz|\rvx)]$ is the distribution of representations for the entire data set, $q(\rvx)$ is the distribution of real data, and $\KL$ is the Kullback-Leibler divergence.

When the total correlation attains the minimum of $0$, we have $q(\rvz)=\prod_j q(\rz_j)$ almost everywhere; that is, factors in the representation are independent. Therefore, penalizing the total correlation simply enforces the model to satisfy the following independence assumption:
% : factors in disentangled representations should be independent. For clarity, we summarize the independence assumption as follows:
\begin{assumption}
	\textbf{(The Marginal Independence Assumption)} \label{ass:common}
	Suppose that the representation $\rvz\in \R^J$ of data variable $\rvx$ are disentangled. Then factors in the representation are independent, i.e.
	\begin{equation}
	q(\rvz)=\prod_{j=1}^J q(\rz_j)
	\end{equation}
	almost everywhere.
\end{assumption}

As discussed in previous sections, independence is an intrinsic property of representations, but disentanglement should also emphasize the relation between data and representations. Therefore, the independence assumption only describes partial properties of disentanglement. 
% Motivated by this, we aim at proposing another assumption to describe the relation between data and disentangled representations.
In the next, we will formulate our additional assumption for disentangled representations mathematically.

\subsection{Proposed New Assumption}
% Our key insight is that we can establish an assumption via mutual information to describe the relation between data and disentangled representations. According to Bengio's statement~\cite{bengio2013representation}, each factor in disentangled representations corresponds to a single factor of variation in data, and is invariant to other factors of variation. Though there is usually no explicit expression for factors of variation, we can characterize the factors of variation in data using mutual information: \textit{the mutual information
% between data and each factor in disentangled representations is invariant to other factors}. This assumption is intuitively necessary but not sufficient for disentanglement, as it cannot ensure the bijective mapping between factors in representations and factors of variation in data.

To formulate our proposed assumption, we begin with introducing mutual information and conditional mutual information. Mutual information measures the information of a variable contained by another, which can be expressed as follows:
\begin{equation}
I(\rvx;\rvz)=H(\rvz)-H(\rvz|\rvx)
\end{equation}
where $H(\rvz)=-\E_{q(\rvz)}[\log q(\rvz)]$ is the entropy and  $H(\rvz|\rvx)=-\E_{q(\rvz,\rvx)}[\log q(\rvz|\rvx)]$ is the conditional entropy. 
%are measures of uncertainty. 
% Hence mutual information is the uncertainty reduction of one variable when another is given.

The conditional mutual information measures the mutual information between two variables given the presence of other variables:
\begin{equation}
I(\rz_j;\rvx|\{\rz_i\}_{i\in S_{-j}})=H(\rz_j|\{\rz_i\}_{i\in S_{-j}})-H(\rz_j|\rvx,\{\rz_i\}_{i\in S_{-j}})
\end{equation}
where $j\in [J]$, $S_{-j}$ is any subset of $[J]\backslash\{j\}$ and $\{\rz_i\}_{i\in S_{-j}}$ denotes all factors with subscript index in $S_{-j}$. Hence the conditional mutual information is the difference of two conditional entropy. 

Using the concepts above, we can elegantly formulate the proposed assumption into mathematical equations:
\begin{assumption} 
	\textbf{(The Proposed Assumption)} \label{ass:proposed1}
	Suppose that the representation $\rvz\in \R^J$ of data $\rvx$ is disentangled. Then for any single factor $\rz_j$, its mutual information with data is invariant to other factors $\{\rz_i\}_{i\in S_{-j}}$, i.e.
	\begin{equation}
	I(\rz_j;\rvx|\{\rz_i\}_{i\in S_{-j}})=I(\rz_j;\rvx) \label{eq:assumption1}
	\end{equation}
\end{assumption}

Note that the conditional mutual information is closely related the chain rule of mutual information. Using chain rule we can derive another equivalent equation for the proposed assumption. First, we have the following lemma:
\begin{lemma}
	$I(\rz_j;\rvx|\{\rz_i\}_{i\in S_{-j}})=I(\rz_j;\rvx)$ for any $j$ and $S_{-j}$ is equivalent to the following equation:
	\begin{equation}
	I(\{\rz_i\}_{i\in S};\rvx)=I(\rz_j;\rvx)+I(\{\rz_i\}_{i\in S_{-j}};\rvx)
	\label{eq:info_decomposition}
	\end{equation} 
	where $S=\{j\}\cup S_{-j}$ is any subset of $[J]$, $j$ is any single element in $S$.
\end{lemma}

Iteratively using Eq.~\ref{eq:info_decomposition}, we can formulate the proposed assumption into another equivalent equation:
\begin{assumption}
	\textbf{(Reformulation of the above Proposed Assumption)} \label{ass:proposed2}
	Suppose that the representation $\rvz\in \R^J$ of data  $\rvx$ is disentangled. Then for any subset of factors in the representation, we have
	\begin{equation}
	I(\{\rz_i\}_{i\in S};\rvx)=\sum_{i\in S} I(\rz_i;\rvx)\,, \label{eq:assumption2}
	\end{equation}
	where $S\subset [J]$.
\end{assumption}
This equation is more suitable for further analysis, as it avoids conditional mutual information and hence becomes cleaner and easier to deal with.

At the end, we want readers to note that $I(\rvz;\rvx)$ is not finite when $\rvz$ and $\rvx$ are continuously distributed and the relation between $\rvz$ and $\rvx$ is deterministic and bijective. 
This will not bring us big troubles since most models we are studying are free from this problem, and deterministic representations can be studied through stochastic ones by adding shrinking continuous noise.
%highlight that the scope of application of the proposed assumption above is limited in models estimating the posterior $q(\rvz|\rvx)$, i.e. the latent variable $\rvz$ is stochastic when data $\rvx$ is given. Otherwise, if there exists a deterministic transformation from $\rvx$ to $\rvz$, i.e. $q(\rvz|\rvx)$ vanishes into a delta distribution, then the mutual information $I(\rvx;\rvz)$ is positively infinite and hence is not well-defined (\textbf{proof deferred to the supplemental materials}).

\section{Conditional Independence and Factorized Approximate Posterior}
In this section, we first connect our proposed assumption with conditional independence of $\rvz$ conditioned on $\rvx$. This reveals the necessity of such conditional independence for disentanglement. Then we show that conditional independence of factors has already been satisfied by the encoders of VAEs and InfoGAN due to the use of factorized approximate posterior, which we believe is an important inductive bias for models to learn disentangled representations. Finally, we show that the conditional independence of factors can be attributed to a factorized noise in the form of reparameterization~\cite{kingma2014auto-encoding}, which is related to the design of our experiments.

\subsection{Connection with Conditional Independence} \label{sec:connection}
Mutual information in Eq.~\ref{eq:assumption2} is still a hard quantity to evaluate. To see its implications on model design, we need to further simplify it. For this purpose, we rewrite Eq.~\ref{eq:assumption2} to an equivalent form $I(\{\rz_i\}_{i\in S};\rvx)-\sum_{i\in S} I(\rz_i;\rvx)=0$. The difference on the left hand side can be reduced to quantities involving $\rvz$'s marginal distribution and the $\rvz$'s distribution conditioned on data $\rvx$, as shown by the following theorem:
\begin{theorem}
	For any subset $S$ of $[J]$, we have:
	\begin{align}
	\begin{split}
	 & I(\{\rz_i\}_{i\in S};\rvx)-\sum_{i\in S}I(\rz_i;\rvx) \\
	 & \qquad =\E_{q(\rvx)}\left[\KL\left(q(\{\rz_i\}_{i\in S}|\rvx)\Vert \prod_{i\in S}q(\rz_i|\rvx)\right)\right]
	-\KL\left(q(\{\rz_i\}_{i\in S})\Vert \prod_{i\in S}q(\rz_i)\right) \label{eq:relation}
	\end{split}
	\end{align}
\end{theorem}

Note that the first term on the right hand side is the total correlation of the conditional distribution of $\rvz$ on $\rvx$, and the second term is the total correlation of the marginal distribution of $\rvz$. This theorem bridges disentanglement with conditional independence and marginal independence of factors in representations. 
Since the independence among factors of $\rvz$, that is the factorized $q(\rvz)$, is widely accepted as a necessary condition for disentangled representations, the second term on the right hand side is just $0$ for disentangled representations. And hence for $I(\{\rz_i\}_{i\in S};\rvx)-\sum_{i\in S} I(\rz_i;\rvx)$ to be zero, we only need the first term to be zero.  
%As the proposed assumption pointed out, Eq.~\ref{eq:assumption2} is necessary for disentanglement, while $I(\rvz_{i\in S};\rvx)-\sum_{i\in S} I(\rz_i;\rvx)$ can be decomposed into the difference of two KL divengences: the first KL divergence is about conditional independence, as it reaches zero whenever $\rvz$ is independent conditioned on $\rvx$; and the second KL divergence is about independence, as it reaches zero whenever $\rvz$ is independent.
Therefore, the proposed assumption reveals that the conditional independence (conditioned on data) of factors in representations is also necessary for disentanglement in additional to the marginal independence. %According to the independence assumption, independence of factors is necessary for disentanglement~\cite{kim2018disentangling,chen2018isolating}. Based on this, the proposed assumption holds if and only if the first KL divergence is zero almost everywhere, i.e. conditional independence is satisfied.

\subsection{Inductive Bias: Factorized Approximate Posterior} \label{sec:inductive}
The implication of Eq.~\ref{eq:relation} is quite surprising. People who are familiar with VAE-type models will immediately tell that the conditional independence is automatically satisfied when the factorized structure is chosen for $q(\rvz|\rvx)$. Though such a factorized structure is ubiquitous in variants of VAE, we do not find any clues from literature connecting it with disentanglement. It is most likely just a convenient choice inherited from Kingma et al.'s work where VAE was proposed for the first time\cite{kingma2014auto-encoding}. Such an arrangement also appears in InfoGAN~\cite{chen2016infogan}. 
%investigating inductive biases which ensures conditional independence of factors in representations. Specifically, we point out that the factorized approximate posterior $q(\rvz|\rvx)$ ensures conditional independence, and hence is exactly an inductive bias for disentanglement in VAEs~\cite{kingma2014auto-encoding} and InfoGAN~\cite{chen2016infogan}.

%In the field of disentangled representation learning, models estimating posterior mainly includes VAEs~\cite{kingma2014auto-encoding} and InfoGAN~\cite{chen2016infogan}. These models usually estimate the posterior by Gaussian distribution $\mathcal{N}(\rvz|\mu(\vx),\sigma^2(\vx)\mI)$, where the covariance matrix is diagonal and hence the approximate posterior $q(\rvz|\rvx)$ satisfies conditional independence. Combined with the proposed assumption, we can conclude that factorized approximate posterior is an inductive bias for disentanglement in VAEs and InfoGAN.

Equation~\ref{eq:relation} implies the proposed assumption only when the marginal independence also holds. Though the marginal independence of representations is arguably achievable in practice, it is always part of the objectives of these representation learning algorithms, sometimes implicitly. In particular, for VAE-type models, when the conditional independence is hard coded in the encoders, enhancing the marginal independence can further improve disentanglement of the representations. On the other hand, InfoGAN's decoder part satisfies the marginal independence, and penalizing the information term with factorized approximate posterior can enhance the conditional independence in decoder and thus improve disentanglement of the generated samples.
Considering that the marginal independence has been emphasized in many previous works, and it alone cannot guarantee the disentanglement, we believe that the conditional independence is the more important but overlooked inductive bias for the disentanglement in the representation learning algorithms.

% \subsection{Reformulating Conditional Independence}
Using the form of reparameteization~\cite{kingma2014auto-encoding}, 
data $\vx$ is encoded into a mean vector $\mu(\vx)$ and a variance vector $\sigma(\vx)$, and then the latent variables $\rvz$ is formulated by $\mu(\vx)$, $\sigma(\vx)$ and a noise $\epsilon$ as follows:
\begin{equation}
    \rvz=\mu(\vx)+\sigma(\vx)\odot\epsilon\,, \epsilon\sim \mathcal{N}(0, \mI)\,.
\end{equation}
It is easy to show that the conditional independence of factors of $\rvz$ on data can be attributed to the independence of noise, as we summarize in Lemma~\ref{lem:cond_ind}. By controlling the covariance structure of the noise vector, we can control the level of compliance of models with the conditional independence in our experiments.

%Note that the noise $\epsilon$ follows standard Gaussian distribution $\mathcal{N}(0, \mI)$, thus it is factorized, i.e. $q_{\epsilon}(\epsilon)=\prod_{j=1}^J q_{\epsilon_j}(\epsilon_j)$. Then the form of the approximate posterior $q(\rvz|\vx)$ is determined by the form of noise distribution $q_{\epsilon}(\epsilon)$ (\textbf{proof deferred to the supplemental materials}):
 \begin{lemma}~\label{lem:cond_ind}
     Suppose $\rvz=\mu(\rvx)+\sigma(\rvx)\odot\epsilon$, $\epsilon\sim q_{\epsilon}(\epsilon)$ and $\sigma_j(\rvx)>0$ for any $j\in [J]$, then the independence of $\rvz$ conditioned on $\rvx$ is equivalent to the independence of $\epsilon$:
     \begin{equation}
         q(\rvz|\rvx)=\prod_{j=1}^Jq(\rz_j|\rvx)\Leftrightarrow q_{\epsilon}(\epsilon)=\prod_{j=1}^J q_{\epsilon_j}(\epsilon_j)
     \end{equation}
 \end{lemma}

% \begin{equation}
%     q(\rvz|\vx)=\frac{1}{\prod_{j=1}^J\sigma_j(\vx)}q_{\epsilon}(\frac{\rz_1-\mu_1(\vx)}{\sigma_1(\vx)}, \dots, \frac{\rz_J-\mu_J(\vx)}{\sigma_J(\vx)})=\prod_{j=1}^J\frac{1}{\sigma_j(\vx)}q_{\epsilon_j}(\frac{\rz_j-\mu_j(\vx)}{\sigma_j(\vx)})
% \end{equation}

% We can see from the equation above that the independence of latent variables $\rvz$ conditioned on $\vx$ is determined by the independence of noise. Specifically, if and only if $q_{\epsilon}(\epsilon)=\prod_{j=1}^J q_{\epsilon_j}(\epsilon_j)$, the approximate posterior $q(\rvz|\vx)$ can be formulated into the product of the distribution of its factors (see the last formula of the equation above, actually $q(\rz_j|\vx)=\frac{1}{\sigma_j(\vx)}q_{\epsilon_j}(\frac{\rz_j-\mu_j(\vx)}{\sigma_j(\vx)})$). Therefore, conditional independence in reparameterizarion is determined by the factorized noise.

\section{Experiments}
In this section we empirically show the importance of the proposed assumption for disentangled representation learning by investigating the effect of violating the conditional independence in VAEs and InfoGAN.

Facilitated by the Lemma~\ref{lem:cond_ind}, we can use correlated noise in VAEs to violate the conditional independence. Similarly, in InfoGAN we can use the correlated Gaussian distribution as the approximate posterior to violate the conditional independence. 
%These settings do not involve extra parameters, enabling us to fairly compare the experimental results with and without conditional independence.

Specifically, we introduce a correlated noise as follows:
\begin{equation}
\epsilon\sim\mathcal{N}(0, \Sigma), \Sigma=(1-\sigma)\mI+\sigma \1\1^\top
\end{equation}
where $\1$ is a column vector of all $1$'s, and $\sigma\in[0,1)$ is the correlation weight. Factors in noise following Gaussian distribution with covariance matrix $\1\1^\top$ are highly correlated. When $\sigma=0$, the noise is factorized, and larger $\sigma$ corresponds to higher correlation. We compare the impact of setting $\sigma=0.9$ with the baselines, where $\sigma=0$.

Note that the corresponding distribution defined by $\rvz=\mu(\rvx)+\sigma(\rvx)\odot\epsilon$ is a Gaussian as follows:
\begin{equation}\label{eq:corr_pos}
    q(\rvz|\rvx)=\mathcal{N}(\mu(\rvx),\mathrm{diag}(\sigma(\rvx))\Sigma \mathrm{diag}(\sigma(\rvx)))
\end{equation}
where $\mathrm{diag}(\cdot)$ is a function mapping a vector to a diagonal matrix. This distribution has a nice property: $D_{KL}(q(\rvz|\rvx)\Vert\mathcal{N}(0,\mI))=D_{KL}(\mathcal{N}(\mu(\rvx),\mathrm{diag}(\sigma^2(\rvx)))\Vert\mathcal{N}(0,\mI))+c$, where $c$ is a constant. This property enables us to train VAE models without rewriting the KL term. Similarly, for the experiments of InfoGAN, we set Eq.~\ref{eq:corr_pos} as the approximate posterior with $\sigma=0.9$ for comparisons. These setting does not increase the parameters of models, which ensures the fairness of comparisons.

\textbf{Data sets:} According to~\cite{locatello2019Challenging}, the performance of representation learning models on different data sets can be very different. To demonstrate the impact of the conditional independence on VAEs, we choose three distinct data sets: dSprites~\cite{higgins2017beta-vae}, SmallNORB~\cite{lecun2004learning} and Cars3D~\cite{reed2015deep}. DSprites is a set of 737,280 generated images of size 64*64 in black and white, and the images are generated from fives independent latent factors. SmallNORB contains 24,300 image pairs of 50 3D toys produced by two cameras, each image pair is grey-scale in shape of 2*96*96.
Cars3D consists of 199 colorful 3D car models in shape of 128*128*3*24*4. As for InfoGAN, we choose MNIST, a data set of handwritten digits ranging from 0 to 9, containing 60,000 training testing samples, each being a 28 by 28 grey-scale image.

\textbf{Models:} In experiments of VAEs, we select vanilla VAE, FactorVAE and TCVAE as baselines. 
We choose these models for two reasons: (i) Vanilla VAE only weakly penalizes the total correlation~\cite{chen2018isolating}, while FactorVAE and TCVAE can strongly penalize it by adjusting the their hyperparameters; (ii) FactorVAE and TCVAE penalize total correlation with different methods. The consistent results of our experiments on these models are able to support the necessity of the inductive bias of conditional independence. 
%Note that for comparisons in VAEs and InfoGAN, we simply set $\sigma=0.9$ in their approximate posteriors, which does not involve additional parameters and hence ensures the fairness of comparisons.

\textbf{Metrics:} We use Mutual Information Gap (MIG)~\cite{chen2018isolating} to evaluate disentanglement in VAEs for its usefulness and rationality. MIG is defined by first evaluating the normalized mutual information between each factor in representations and each ground truth factor, then computing the gap between the highest two normalized mutual information values along factors in representations, and finally returning the gap averaged along ground truth factors. We also use reconstruction error and KL term in the objective of VAEs, as well as estimates of total correlation (see supplementary material for discussion of total correlation) to measure other effects of violating conditional independence in VAEs. 
%Note that for fair comparison, we use $D_{KL}(\mathcal{N}(\mu(\rvx),diag(\sigma^2(\rvx)))\Vert\mathcal{N}(0,\mI))$ as the KL metric in all models. 
As for InfoGAN, we simply visualize the generated images and compare the disentanglement qualitatively.

\textbf{Hyper parameters:} For fair comparison, we use the implementation of disentanglement lib introduced by~\cite{locatello2019Challenging} without tuning. The number of factors in representations are set as 10 or 20 (see supplementaries for results of 20 factors). The weights of penalties in FactorVAE and TCVAE are set as 35 and 6 on all experiments, respectively. For each VAE model on each data set, we train it ten times without tuning, and record the MIG scores, reconstruction errors, KL terms and total correlations. While in InfoGAN, we use 2 factors, and the penalizing weight is set to be 0.1.
\begin{figure}[tb!]
%\vspace{-20pt}
%\setlength{\belowcaptionskip}{-0.7cm}
\centering
    \subfigure[dSprites]{
	\includegraphics[width=4.5cm,height=3.5cm]{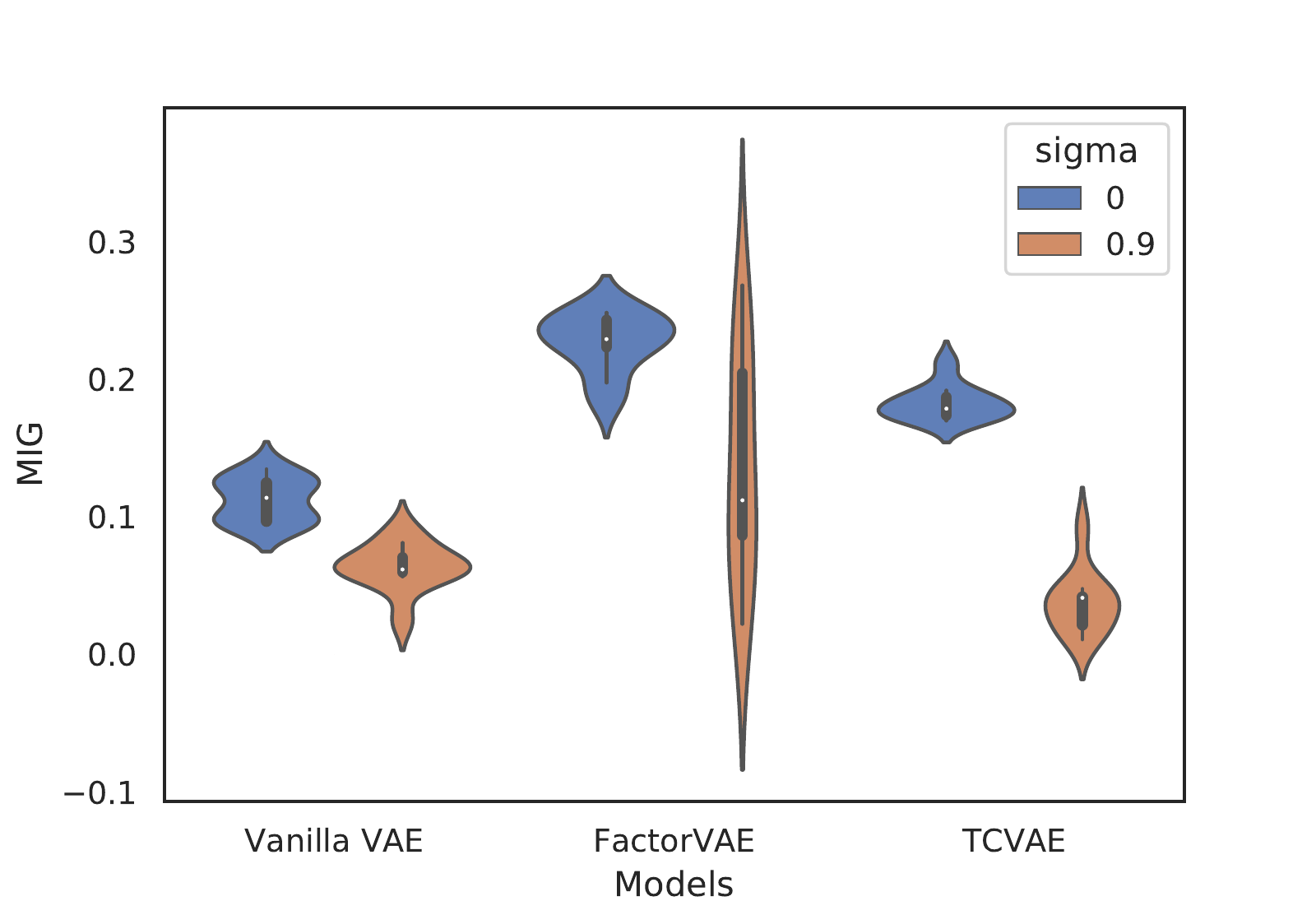}}
	\subfigure[SmallNORR]{
	\includegraphics[width=4.5cm,height=3.5cm]{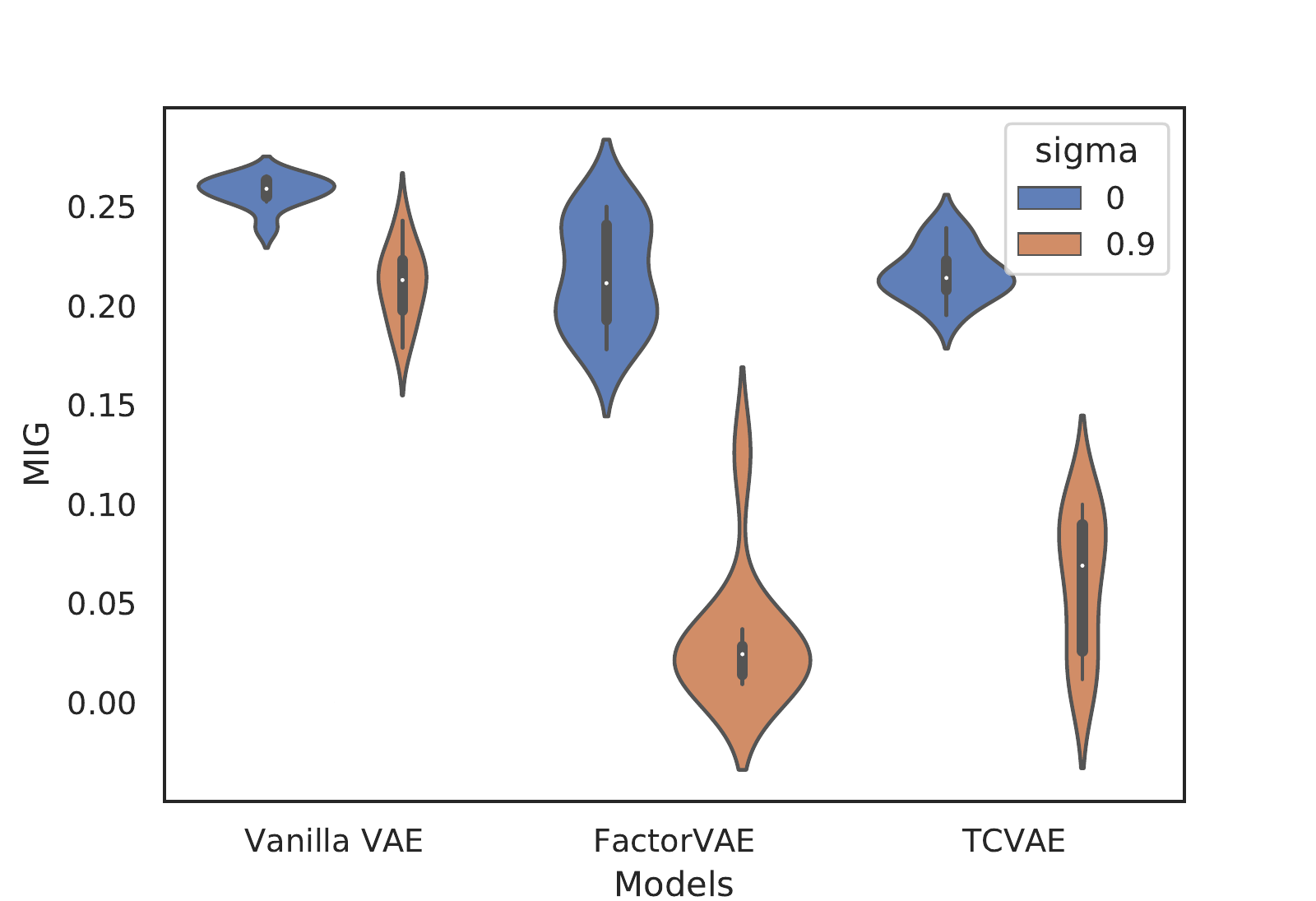}}
	\subfigure[Cars3D]{
	\includegraphics[width=4.5cm,height=3.5cm]{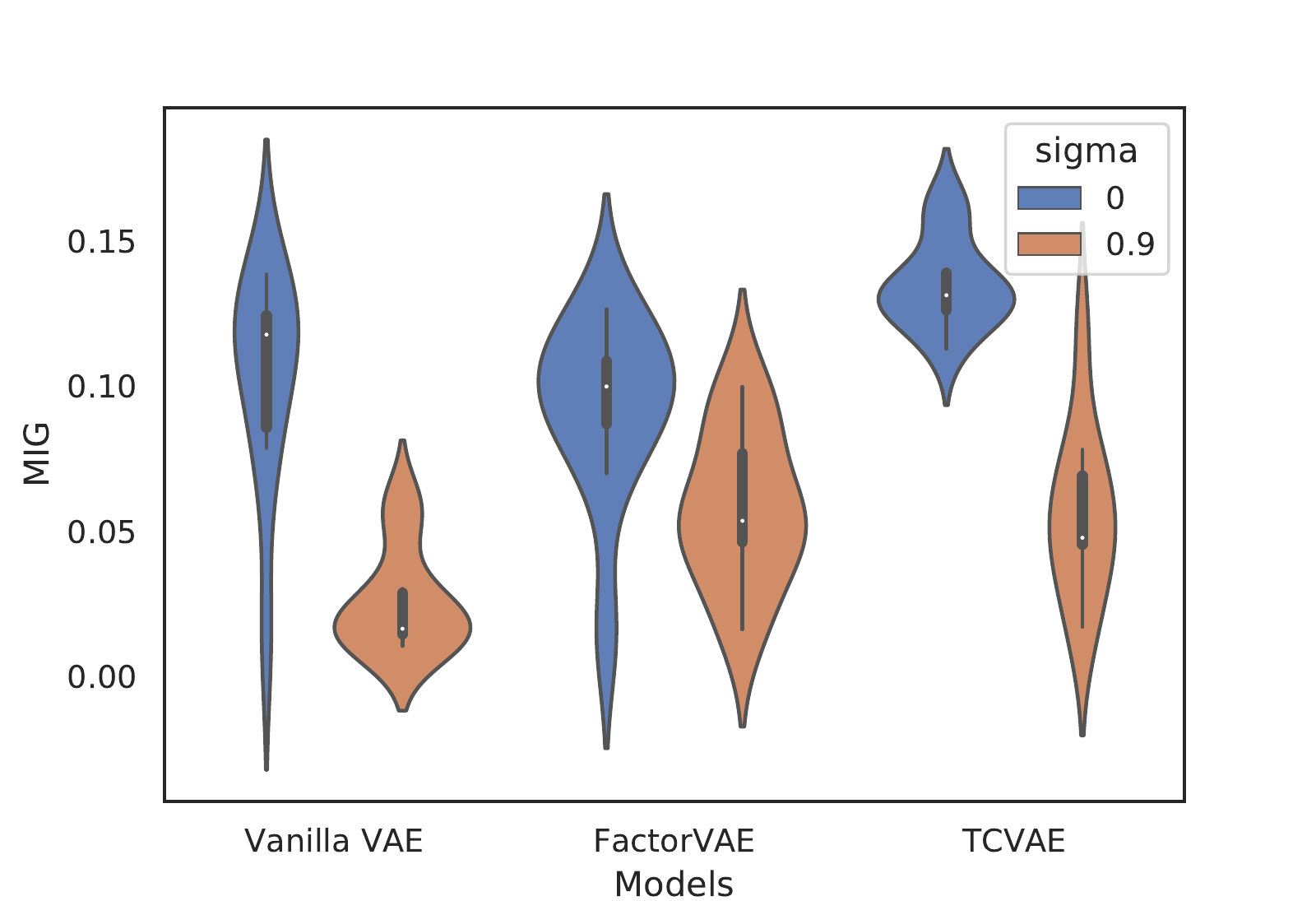}}
	\vspace{-10pt}
	\caption{\textbf{MIG of vanilla VAE, FactorVAE and TCVAE on dSprites, SmallNORR and Cars3D.} Blue and orange violin graphs show MIG distributions when $\sigma=0$ and $\sigma=0.9$, respectively.}
	\label{fig:MIG_J=10}
\end{figure}
\begin{figure}[tb!]
\centering
    \subfigure[Reconstruction Error]{
	\includegraphics[width=4.5cm,height=3.5cm]{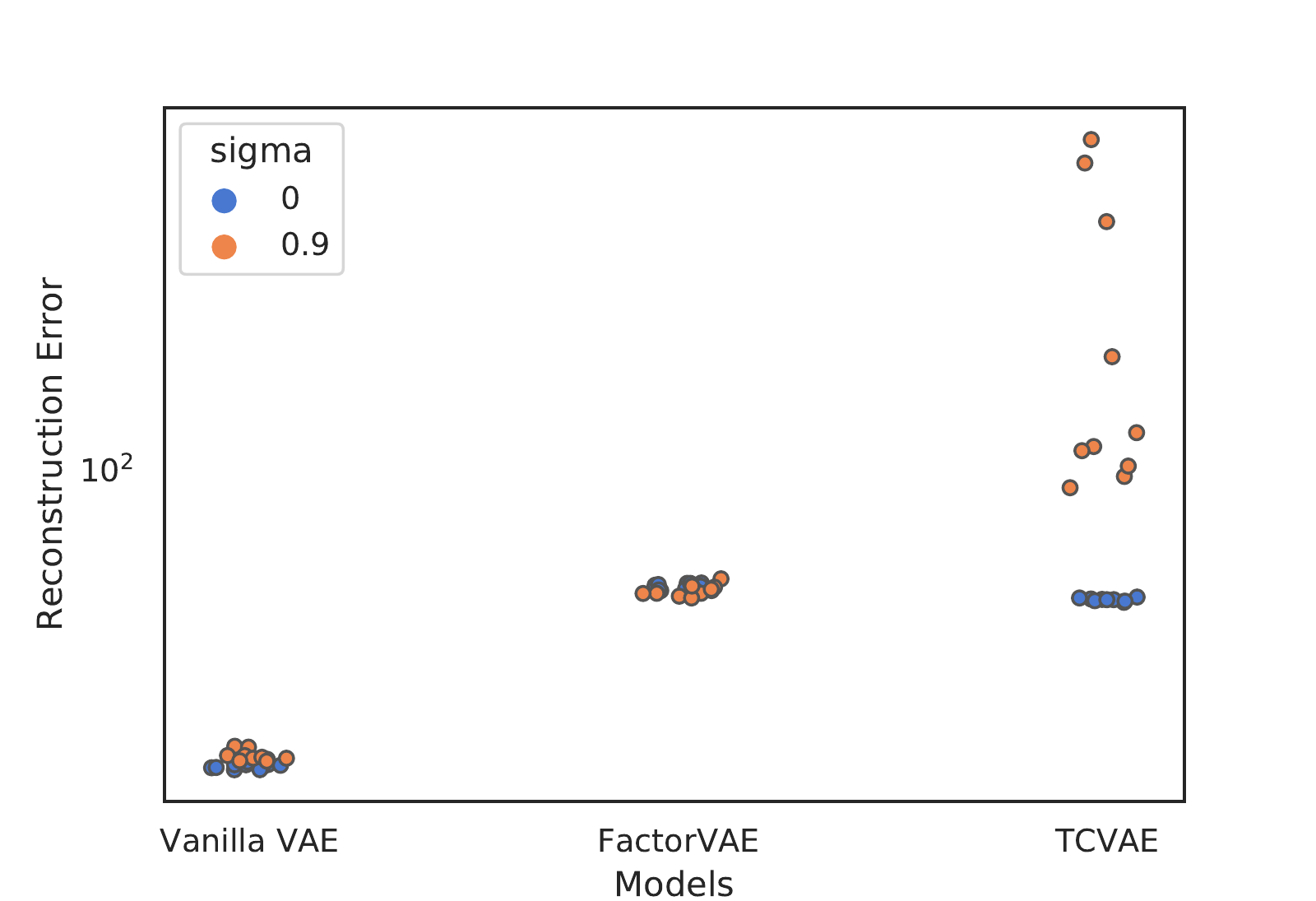}}
	\subfigure[KL Term]{
	\includegraphics[width=4.5cm,height=3.5cm]{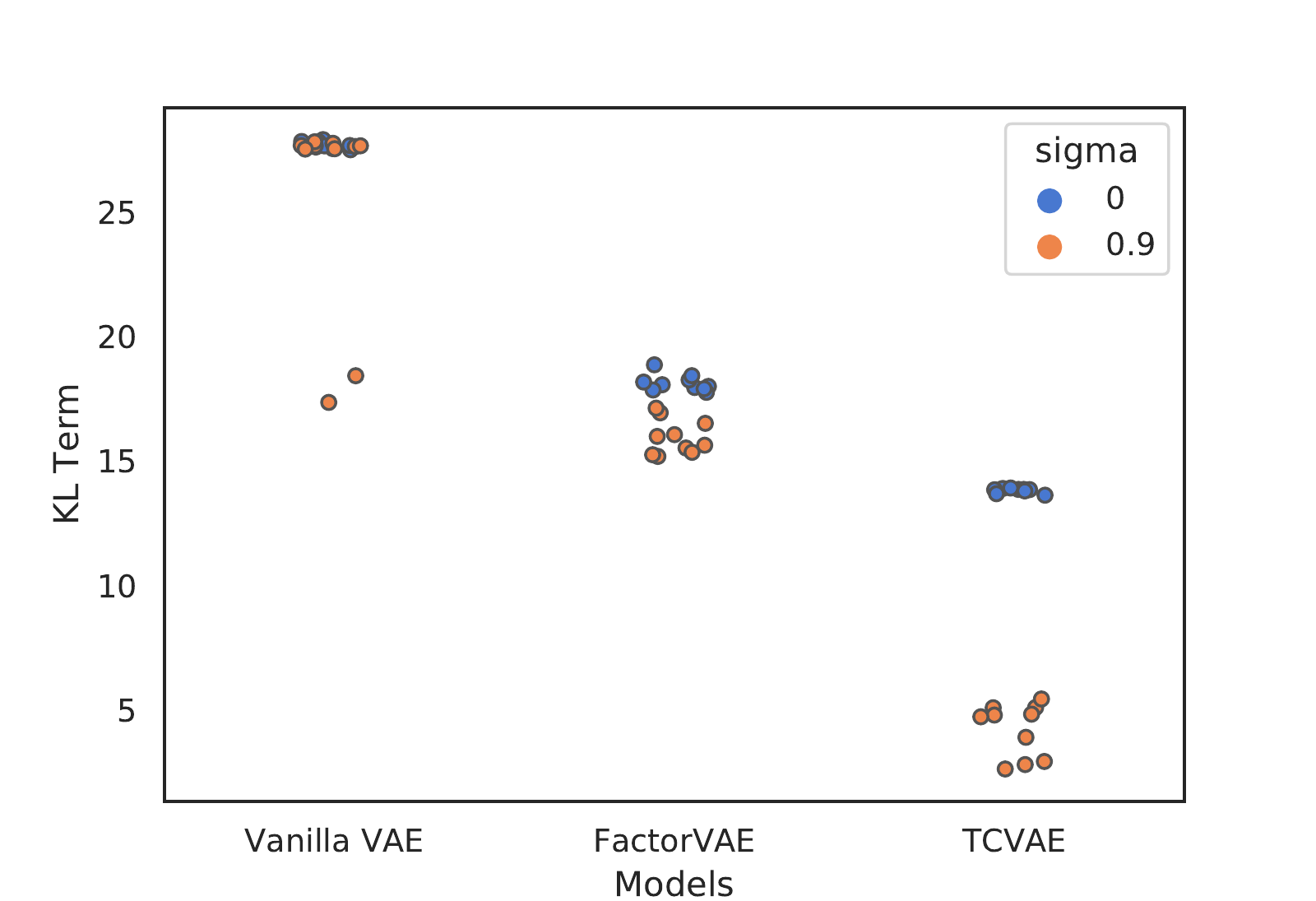}}
	\subfigure[Total Correlation]{
	\includegraphics[width=4.5cm,height=3.5cm]{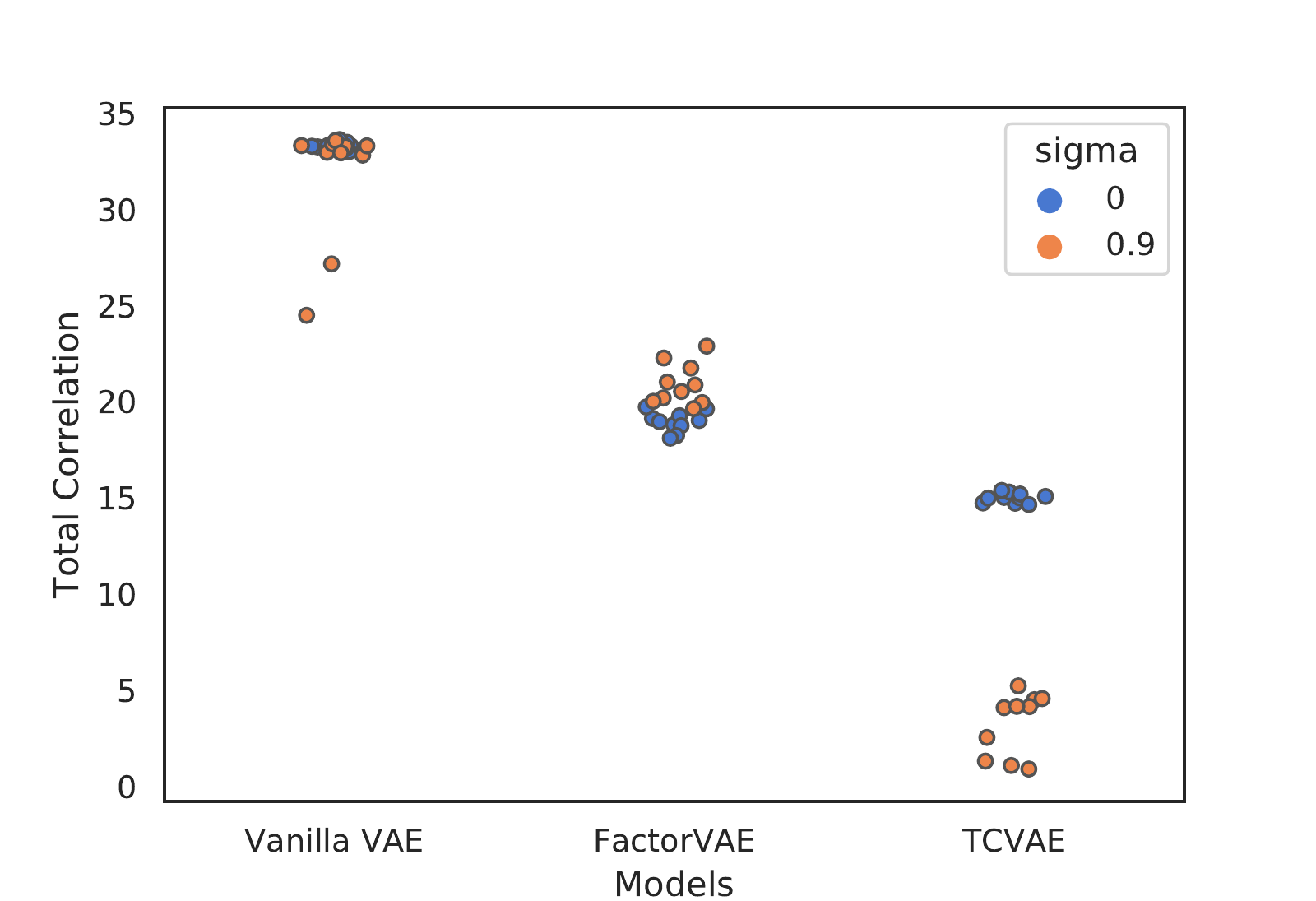}}
	\vspace{-10pt}
	\caption{\textbf{Reconstruction Error, KL Term and Total Correlation of vanilla VAE, FactorVAE and TCVAE on dSprites.} Blue and orange points denote results with $\sigma=0$ and $0.9$, respectively.}
	\label{fig:Metrics_J=10}
\end{figure}

\subsection{Experimental Results on VAEs}
First, we show the decline of disentanglement measured by MIG for vanilla VAE, FactorVAE and TCVAE when using correlated approximate posteriors instead of factorized approximate posteriors in Figure~\ref{fig:MIG_J=10}. The declining of MIG scores is consistent across different models and different data sets when $\sigma$ varies from $0$ to $0.9$. These results reveal that factorized approximate posterior is important for these disentangled representation learning models. 

Moreover, we also consider the impact of using correlated approximate posterior in VAEs on reconstruction error, KL term and total correlation. As shown in Figure~\ref{fig:Metrics_J=10}, these metrics are nearly unchanged for vanilla VAE and FactorVAE on dSprites when $\sigma$ varies from $0$ to $0.9$. While the case of TCVAE is different, it has higher reconstruction errors but lower KL terms and total correlations when $\sigma=0.9$, which might be due to the trade-off between reconstruction and the KL term. 
%Note that the total correlation reflects the independence of factors, which is independent of conditional independence. 
Overall, we can conclude that the correlated approximate posterior only causes the decline of disentanglement without strong impact on  reconstruction errors, KL terms or total correlations.

To conclude, the results of VAEs with $\sigma=0$ and $\sigma=0.9$ show that the factorized approximate posterior is important for disentanglement of representations learned by VAE-based models. This undoubtly supports the necessity of our proposed assumption for disentanglement.
\begin{figure}[tb!]
\centering
	\subfigure[Digit 4 ($\sigma=0$)]{
	\includegraphics[width=3.3cm,height=3.3cm]{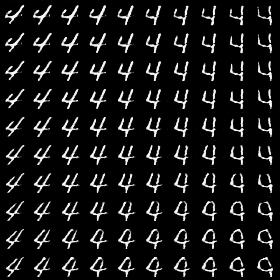}}
	\subfigure[Digit 7 ($\sigma=0$)]{
	\includegraphics[width=3.3cm,height=3.3cm]{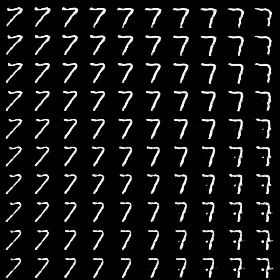}}
	\subfigure[Digit 4 ($\sigma=0.9$)]{
	\includegraphics[width=3.3cm,height=3.3cm]{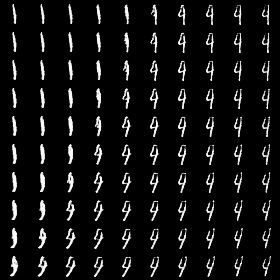}}
	\subfigure[Digit 7 ($\sigma=0.9$)]{
	\includegraphics[width=3.3cm,height=3.3cm]{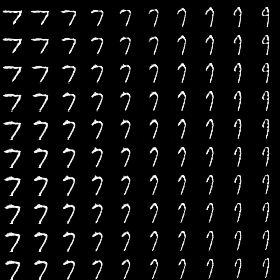}}
	\vspace{-5pt}
	\caption{\textbf{Varying continuous factors in InfoGAN with $\sigma=0$ and $\sigma=0.9$ on MNIST.} The horizontal and the vertical axes represent varying the first and the second factors on $[-2,2]$, respectively.}
	\label{fig:InfoGAN_varying}
\end{figure}
\subsection{Experimental Results on InfoGAN}
We show in Figure~\ref{fig:InfoGAN_varying} the decline of disentanglement in InfoGAN when using correlated approximate posterior for factors. We traverse the first and the second continuous factors in InfoGAN's decoder with other factors fixed, and then compare the generated images. We do not use the MIG score of encoder's outputs as the independence of factors is not pursued in encoder like VAEs. The images above clearly show that violating conditional independence leads to decline of disentanglement of InfoGAN. In the case of $\sigma=0$, the horizontal and the vertical axes correspond to the rotation and width of generated digits. While in the case of $\sigma=0.9$, no obvious factor of variation in MNIST can be identified or bond to the two factors we visualize. And varying continuous factors leads to changing of digits. These results once again demonstrate the importance of the factorized approximate posterior in InfoGAN for disentanglement.

\section{Conclusion}
We propose an assumption characterizing disentangled representations through the interaction between the factors of representations and data in terms of mutual information: the mutual information between data and each factor in disentangled representations is invariant with respect to the other factors. We bridge it with independence of factors conditioned on data, and then point out that the factorized approximate posterior is a key inductive bias in the representation learning models for disentanglement. Our experiments of replacing factorized approximate posteriors by correlated ones in various representation learning models support our analysis and demonstrate the importance of our proposed assumption.
It will be important to explore the relation between the necessary conditions we studied here and the sufficient conditions proposed based on nonlinear ICA theory in the future. 
%, without harming the fairness of comparisons. 
% The experimental results convincingly show that our proposed assumption and conditional independence is essential to disentanglement.

\clearpage
\newpage
\section*{Broader Impact}
Disentangled representation learning has been an active area in machine learning recently, which aims to find more effective, interpretable and robust representations towards data. Such feature extraction can be across images, texts, speech etc, though the experiments in this paper are mainly focused on image data. In consequence it may enhance some AI system and application like face recognition, which in turn can cause privacy issues for the society. This calls for more efforts for privacy protection.

% while its usefulness can be readily extended to other domains

%Authors are required to include a statement of the broader impact of their work, including its ethical aspects and future societal consequences.
%Authors should discuss both positive and negative outcomes, if any. For instance, authors should discuss a)
%who may benefit from this research, b) who may be put at disadvantage from this research, c) what are the consequences of failure of the system, and d) whether the task/method leverages biases in the data. If authors believe this is not applicable to them, authors can simply state this.

%Use unnumbered first level headings for this section, which should go at the end of the paper. {\bf Note that this section does not count towards the eight pages of content that are allowed.}

%\clearpage
%\newpage
{\small
\bibliographystyle{ieee_fullname}
\bibliography{egbib}
}

\clearpage
\newpage

\section*{\centering{Supplementary Material}}
\subsection*{A$\quad$ Proofs of Lemmas and Theorems}
\noindent\textbf{Lemma 1.} \textit{$I(\rz_j;\rvx|\{\rz_i\}_{i\in S_{-j}})=I(\rz_j;\rvx)$ for any $j$ and $S_{-j}$ is equivalent to the following equation:
\begin{equation*}
	I(\{\rz_i\}_{i\in S};\rvx)=I(\rz_j;\rvx)+I(\{\rz_i\}_{i\in S_{-j}};\rvx)
\end{equation*} 
where $S=\{j\}\cup S_{-j}$ is any subset of $[J]$, $j$ is any single element in $S$.}\\
\textit{proof.} This lemma is a straight corollary of chain rule of mutual information. To be self contained, here we derive it from scratch using Bayes rule:
\begin{equation*}
\begin{split}
    &I(\rz_j;\rvx|\{\rz_i\}_{i\in S_{-j}})\\
    =&\E_{q(\{\rz_i\}_{i\in S},\rvx)}\left[\log\frac{q(\rz_j,\rvx|\{\rz_i\}_{i\in S_{-j}})}{q(\rz_j|\{\rz_i\}_{i\in S_{-j}})q(\rvx|\{\rz_i\}_{i\in S_{-j}})}\right]\\
    =&\E_{q(\{\rz_i\}_{i\in S},\rvx)}\left[\log\frac{q(\rvx|\{\rz_i\}_{i\in S})}{q(\rvx|\{\rz_i\}_{i\in S_{-j}})}\right]\\
    =&\E_{q(\{\rz_i\}_{i\in S},\rvx)}\left[\log\frac{q(\rvx,\{\rz_i\}_{i\in S})q(\{\rz_i\}_{i\in S_{-j}})}{q(\{\rz_i\}_{i\in S})q(\rvx,\{\rz_i\}_{i\in S_{-j}})}\right]\\
    =&\E_{q(\{\rz_i\}_{i\in S},\rvx)}\left[\log\frac{q(\rvx,\rvz_{i\in S})}{q(\{\rz_i\}_{i\in S})q(\rvx)}-\log\frac{q(\rvx,\{\rz_i\}_{i\in S_{-j}})}{q(\{\rz_i\}_{i\in S_{-j}})q(\rvx)}\right]\\
    =&I(\{\rz_i\}_{i\in S};\rvx)-I(\{\rz_i\}_{i\in S_{-j}};\rvx)
\end{split}
\end{equation*}
Hence we have:
\begin{equation*}
\begin{split}
    I(\rz_j;\rvx|\{\rz_i\}_{i\in S_{-j}})=I(\rz_j;\rvx)
    \Leftrightarrow I(\{\rz_i\}_{i\in S};\rvx)=I(\rz_j;\rvx)+I(\{\rz_i\}_{i\in S_{-j}};\rvx)
\end{split}
\end{equation*}

\noindent\textbf{Theorem 1.} \textit{For any subset $S$ of $[J]$, we have:
	\begin{align*}
	\begin{split}
	 & I(\{\rz_i\}_{i\in S};\rvx)-\sum_{i\in S}I(\rz_i;\rvx)\\
	 & \qquad =\E_{q(\rvx)}\left[\KL\left(q(\{\rz_i\}_{i\in S}|\rvx)\Vert \prod_{i\in S}q(\rz_i|\rvx)\right)\right]
	-\KL\left(q(\{\rz_i\}_{i\in S})\Vert \prod_{i\in S}q(\rz_i)\right) 
	\end{split}
	\end{align*}}
\textit{proof.} 
\begin{equation*}
\begin{split}
    &I(\{\rz_i\}_{i\in S};\rvx)-\sum_{i\in S} I(\rz_i;\rvx)\\
    =&\E_{q(\{\rz_i\}_{i\in S},\rvx)}\left[\log\frac{q(\{\rz_i\}_{i\in S},\rvx)}{q(\{\rz_i\}_{i\in S})q(\rvx)}\right]-\sum_{i\in S}\E_{q(\rz_i,\rvx)}\left[\log\frac{q(\rz_i,\rvx)}{q(\rz_i)q(\rvx)}\right]\\
    =&\E_{q(\{\rz_i\}_{i\in S},\rvx)}\left[\log\frac{q(\{\rz_i\}_{i\in S},\rvx)}{q(\{\rz_i\}_{i\in S})q(\rvx)}-\sum_{i\in S}\log\frac{q(\rz_i,\rvx)}{q(\rz_i)q(\rvx)}\right]\\
    =&\E_{q(\{\rz_i\}_{i\in S},\rvx)}\left[\log\frac{q(\{\rz_i\}_{i\in S}|\rvx)}{q(\{\rz_i\}_{i\in S})}-\sum_{i\in S}\log\frac{q(\rz_i|\rvx)}{q(\rz_i)}\right]\\
    =&\E_{q(\{\rz_i\}_{i\in S},\rvx)}\left[\log\frac{q(\{\rz_i\}_{i\in S}|\rvx)\prod_{i\in S} q(\rz_i)}{q(\{\rz_i\}_{i\in S})\prod_{i\in S} q(\rz_i|\rvx)}\right]\\
    =&\E_{q(\{\rz_i\}_{i\in S},\rvx)}\left[\log\frac{q(\{\rz_i\}_{i\in S}|\rvx)}{\prod_{i\in S} q(\rz_i|\rvx)}-\log\frac{q(\{\rz_i\}_{i\in S})}{\prod_{i\in S} q(\rz_i)}\right]\\
    =&\E_{q(\rvx)}\left[\KL\left(q(\{\rz_i\}_{i\in S}|\rvx)\Vert \prod_{i\in S} q(\rz_i|\rvx)\right)\right] -\KL\left(q(\{\rz_i\}_{i\in S})\Vert \prod_{i\in S} q(\rz_i)\right)
\end{split}
\end{equation*}

\noindent\textbf{Lemma 2.} \textit{Suppose $\rvz=\mu(\rvx)+\sigma(\rvx)\odot\epsilon$, $\epsilon\sim q_{\epsilon}(\epsilon)$ and $\sigma_j(\rvx)>0$ for any $j\in [J]$, then the independence of $\rvz$ conditioned on $\rvx$ is equivalent to the independence of $\epsilon$:
\begin{equation*}
    q(\rvz|\rvx)=\prod_{j=1}^Jq(\rz_j|\rvx)\Leftrightarrow q_{\epsilon}(\epsilon)=\prod_{j=1}^J q_{\epsilon_j}(\epsilon_j)
\end{equation*}}
\textit{proof.} Note that $\rvz=\mu(\rvx)+\sigma(\rvx)\odot\epsilon$ is a linear transformation between $\epsilon$ and $\rvz$, and hence the approximate posterior $q(\rvz|\rvx)$ can be expressed through the form of $q_{\epsilon}(\cdot)$, and vice versa:
\begin{equation*}
\begin{split}
    &q(\rvz|\rvx)=\frac{1}{\prod_{j=1}^J\sigma_j(\rvx)}q_{\epsilon}(\frac{\rz_1-\mu_1(\rvx)}{\sigma_1(\rvx)}, \dots, \frac{\rz_J-\mu_J(\rvx)}{\sigma_J(\rvx)})\\
    &q_{\epsilon}(\epsilon)=(\prod_{j=1}^J\sigma_j(\rvx))q(\mu_1(\rvx)+\epsilon_1\sigma_1(\rvx),\cdots,\mu_J(\rvx)+\epsilon_J\sigma_J(\rvx)|\rvx)
\end{split}
\end{equation*}
where $\frac{1}{\prod_{j=1}^J\sigma_j(\rvx)}$ and $\prod_{j=1}^J\sigma_j(\rvx)$ is determinants of the corresponding Jacobians.\\
If $q_{\epsilon}(\epsilon)=\prod_{j=1}^J q_{\epsilon_j}(\epsilon_j)$, then we have:
\begin{equation*}
    q(\rvz|\rvx)=\prod_{j=1}^J\frac{1}{\sigma_j(\rvx)}q_{\epsilon_j}(\frac{\rz_j-\mu_j(\rvx)}{\sigma_j(\rvx)})=\prod_{j=1}^J q(\rz_j|\rvx)
\end{equation*}
where $q(\rz_j|\rvx)=\frac{1}{\sigma_j(\rvx)}q_{\epsilon_j}(\frac{\rz_j-\mu_j(\rvx)}{\sigma_j(\rvx)})$.\\
On the other hand, if $q(\rvz|\rvx)=\prod_{j=1}^Jq(\rz_j|\rvx)$, similarily we have:
\begin{equation*}
    q_{\epsilon}(\epsilon)=(\prod_{j=1}^J\sigma_j(\rvx))q(\mu_j(\rvx)+\epsilon_j\sigma_j(\rvx)|\rvx)=\prod_{j=1}^J q_{\epsilon_j}(\epsilon_j)
\end{equation*}
where $q_{\epsilon_j}(\epsilon_j)=\sigma_j(\rvx)q(\mu_j(\rvx)+\epsilon_j\sigma_j(\rvx)|\rvx)$.

\subsection*{B$\quad$ Correlated Approximate Posterior}
As shown above, conditional independence of factors of $\rvz$ on data $\rvx$ can be attributed to the independence of noise $\epsilon$. This enables us to control the level of compliance of approximate posterior with the conditional independence by controlling the covariance strucutre of the noise.

Motivated by this, we define correlated approximate posterior as follows:
\begin{equation*}
    \rvz=\mu(\rvx)+\sigma(\rvx)\odot\epsilon, \epsilon\sim\mathcal{N}(0, \Sigma), \Sigma=(1-\sigma)\mI+\sigma \1\1^\top
\end{equation*}
where $\sigma\in[0,1)$ is the correlation weight. When $\sigma=0$, $\epsilon\sim\mathcal{N}(0, \mI)$ and hence the approximate posterior $q(\rvz|\rvx)$ is factorized. When $\sigma>0$, $q(\rvz|\rvx)$ is correlated, and higher $\sigma$ leads to higher correlation.

The forms of $q(\rvz|\rvx)$ can be expressed by $q_\epsilon(\cdot)$, which is a Gaussian density as follows:
\begin{equation*}
\begin{split}
    & q(\rvz|\rvx)=\frac{1}{\prod_{j=1}^J\sigma_j(\rvx)}q_{\epsilon}(\mathrm{diag}(\sigma(\rvx))^{-1}(\rvz-\mu(\rvx)))\\
    =& \frac{\exp\{-\frac{1}{2}[\mathrm{diag}(\sigma(\rvx))^{-1}(\rvz-\mu(\rvx))]^\top\Sigma^{-1}[\mathrm{diag}(\sigma(\rvx))^{-1}(\rvz-\mu(\rvx))]\}}{(2\pi)^{J/2}\det(\Sigma)^{1/2}\prod_{j=1}^J\sigma_j(\rvx)}\\
    =& \frac{\exp\{-\frac{1}{2}(\rvz-\mu(\rvx))^\top[\mathrm{diag}(\sigma(\rvx))\Sigma\mathrm{diag}(\sigma(\rvx))]^{-1}(\rvz-\mu(\rvx))\}}{(2\pi)^{J/2}\det(\mathrm{diag}(\sigma(\rvx))\Sigma\mathrm{diag}(\sigma(\rvx)))^{1/2}}\\
    =& \mathcal{N}(\rvz|\mu(\rvx),\mathrm{diag}(\sigma(\rvx))\Sigma\mathrm{diag}(\sigma(\rvx)))
\end{split}
\end{equation*}
where $\det(\cdot)$ is the determinant.

The KL divergence between this density and $\mathcal{N}(0,\mI)$ is:
\begin{equation*}
\begin{split}
    &\KL\left(q(\rvz|\rvx)\Vert\mathcal{N}(0,\mI)\right)\\
    =&\frac{1}{2}\left\{\log\frac{1}{\det(\mathrm{diag}(\sigma(\rvx))\Sigma\mathrm{diag}(\sigma(\rvx)))}-J+\mathrm{tr}(\mathrm{diag}(\sigma(\rvx))\Sigma\mathrm{diag}(\sigma(\rvx)))+\mu(\rvx)^\top\mu(\rvx)\right\}
\end{split}
\end{equation*}
where $\mathrm{tr}(\cdot)$ is the trace. Note that the diagonal entries of $\Sigma$ are all 1 and $\mathrm{diag}(\sigma^2(\rvx))$ is a diagonal matrix, and hence $\det(\mathrm{diag}(\sigma(\rvx))\Sigma\mathrm{diag}(\sigma(\rvx)))=\det(\Sigma)\det(\mathrm{diag}(\sigma^2(\rvx)))$, and $\mathrm{tr}(\mathrm{diag}(\sigma(\rvx))\Sigma\mathrm{diag}(\sigma(\rvx)))=\mathrm{tr}(\Sigma\mathrm{diag}(\sigma^2(\rvx)))=\sum_{i=1}^J\sigma_j^2(\rvx)=\mathrm{tr}(\mathrm{diag}(\sigma^2(\rvx)))$. The KL divergence above, therefore, can be further simplified and then be connected with the KL terms of VAEs $\KL\left(\mathcal{N}(\rvz|\mu(\rvx),\mathrm{diag}(\sigma^2(\rvx)))\Vert\mathcal{N}(0,\mI)\right)$:
\begin{equation*}
\begin{split}
    &\KL(q(\rvz|\rvx)\Vert\mathcal{N}(0,\mI))\\
    =&\frac{1}{2}\left\{\log\frac{1}{\det(\mathrm{diag}(\sigma^2(\rvx)))}-J+\mathrm{tr}(\mathrm{diag}(\sigma^2(\rvx)))+\mu(\rvx)^\top\mu(\rvx)+\log\frac{1}{\det(\Sigma)}\right\}\\
    =&\KL\left(\mathcal{N}(\rvz|\mu(\rvx),\mathrm{diag}(\sigma^2(\rvx)))\Vert\mathcal{N}(0,\mI)\right)+\frac{1}{2}\log\frac{1}{\det(\Sigma)}
\end{split}
\end{equation*}
This result demonstrates that the difference between $\KL(q(\rvz|\rvx)\Vert\mathcal{N}(0,\mI))$ and the KL terms of VAEs $\KL\left(\mathcal{N}(\rvz|\mu(\rvx),\mathrm{diag}(\sigma^2(\rvx)))\Vert\mathcal{N}(0,\mI)\right)$ is solely a constant when $\Sigma$ is fixed. This property of our involved correlated approximate posterior $q(\rvz|\rvx)$ is quite nice, which enables us to involve the correlated approximate posterior into VAEs for comparisons without changing their KL terms (the correlated approximate posterior only changes the reconstruction term in VAEs by involving correlated noise into reparamaterization). Note that this setting does not increase the parameters of models, which ensures the fairness of comparisons.

\subsection*{C$\quad$ Mutual Information in Deterministic Case}
In this section, we aim at proving a property of mutual information: the mutual information $I(\rvz;\rvx)$ is positively infinite when $\rvz$ and $\rvx$ are continuously distributed and the relation between $\rvz$ and $\rvx$ is deterministic. 

First, note that the determinacy can be formulated into delta posterior: $\rvz=f(\rvx) \Leftrightarrow q(\rvz|\rvx)=\delta(\rvz-f(\rvx))$, where $\delta(\cdot)$ is delta distribution, and $f(\cdot)$ is the deterministic mapping from $\rvx$ to $\rvz$. This enables us to discuss the deterministic case from the perspective of information theory. Then the property is summarized and proved as follows:
\begin{lemma}
Suppose $q(\rvz,\rvx)$ is a joint density of two continuously distributed variables $\rvx\in\R^n$ and $\rvz\in\R^J$, and (i) the entropy of $\rvz$ is limited: $-\infty<H(\rvz)<+\infty$, (ii) the support of $q(\rvx)$ has non-zero measure: $|\textbf{supp}(\rvx)|>0$. If $q(\rvz|\rvx)=\delta(\rvz-f(\rvx))$, then their mutual information is positively infinite: 
\begin{equation*}
    I(\rvx;\rvz)=+\infty
\end{equation*}
\end{lemma}
\textit{Proof.}
According to the definition of mutual information, we have:
\begin{equation*}
    I(\rvz;\rvx)=H(\rvz)-H(\rvz|\rvx)
\end{equation*}
As $H(\rvz)$ is limited, we focus on the second term $H(\rvz|\rvx)$. This term is the expectation of the conditional entropy with fixed $\vx$:
\begin{equation*}
    H(\rvz|\rvx)=\int_{R^n}p(\vx)H(\rvz|\vx)d\vx
\end{equation*}
If $q(\rvz|\rvx)=\delta(\rvz-f(\rvx))$, then for any given $\vx$, $H(\rvz|\vx)$ is negatively infinite. To see this, we first involve quantized version of $\rvz$ when $\vx$ is given (this is a classic technique in inforation theory to connect entropy of continuous variable with discrete variable): divide the range of $\rvz$ into bins of length $\Delta$, then let a quantized variable, denoted as $\rvz^\Delta$, follow such a distribution:
\begin{equation*}
     \{\dots,q_i=\int_{i\Delta^J}q(\rvz|\vx)d\rvz,\dots\}
\end{equation*}
where $i\Delta^J$ denotes the $i$-th bin. Due to $q(\rvz|\rvx)=\delta(\rvz-f(\rvx))$, only one entry of $\{\cdots,q_i,\cdots\}$ is equal to $1$, and others are $0$ for any length $\Delta$. Hence the quantized version $\rvz^\Delta$ has zero entropy: $H(\rvz^\Delta|\vx)=0$. 

Moreover, let $\tilde{q}(\vz_i|\vx)=q_i/\Delta, \vz_i\in{i\Delta^J}$, then $H(\rvz^\Delta|\vx)$ can be connected with $H(\rvz|\vx)$ as follows:
\begin{equation*}
\begin{split}
    H(\rvz^\Delta|\vx)&=-\sum_i q_i\log q_i \\
    &=-\sum_i\tilde{q}(\vz_i|\vx)\Delta\log(\tilde{q}(\vz_i|\vx)\Delta)\\
    &=-\sum_i\Delta\tilde{q}(\vz_i|\vx)\log\tilde{q}(\vz_i|\vx)-\sum_i\tilde{q}(\vz_i|\vx)\Delta\log\Delta\\
    &=-\sum_i\Delta\tilde{q}(\vz_i|\vx)\log\tilde{q}(\vz_i|\vx)-\log\Delta
\end{split}
\end{equation*}
Note that when $\Delta\rightarrow 0$, $\tilde{q}(\vz_i|\vx)\rightarrow q(\rvz|\vx)$, and $-\sum_i\Delta\tilde{q}(\vz_i|\vx)\log\tilde{q}(\vz_i|\vx)\rightarrow H(\rvz|\vx)$. As $\log\Delta\rightarrow -\infty$ when $\Delta\rightarrow 0$ and $H(\rvz^\Delta|\vx)=0$, we can conclude that $H(\rvz|\vx)=-\infty$ for any given $\vx$. 

Therefore, $H(\rvz|\rvx)=-\infty$ as the support of $q(\rvx)$ has non-zero measure, and hence $I(\rvx;\rvz)=+\infty$.

\subsection*{D$\quad$ Estimate of Total Correlation}
There are some works to estimate the total correlation of marginal $q(\rvz)$ with different approaches. In FactorVAE~\cite{kim2018disentangling}, a discriminator is used to estimated the total correlation, but the estimated total correlation underestimate the true value and hence is not suitable for comparisons. The estimate in TCVAE~\cite{chen2018isolating} is also not suitable as it is biased.~\citet{locatello2019Challenging} propose to first encode a mini-batch of data points into representations, and then fit a Gaussian for them, and finally calculate the total correlation of the fitted Gaussian as the estimate. This approach is quite good but it is suitable as a metric only if $q(\rvz)$ is nearly a Gaussian.

In this works, we estimate the total correlation by Monte Carlo directly. Although this approach needs higher cost of computation, it precisely estimate the total correlation, and higher cost is acceptable for calculating a metric. 

Specifically, we first randomly sample a batch from the data set $\{\vx^{(i)}\}_{i=1}^B$, where $B$ is the batch size. Then we encode the batch into a set of approximate posteriors $\{q(\rvz|\vx^{(i)})\}_{i=1}^B$ by the trained encoders, where $q(\rvz|\vx^{(i)})=\mathcal{N}(\rvz|\mu(\vx^{(i)}),\mathrm{diag}(\sigma^2(\vx^{(i)})))$. Using these approximate posteriors, the marginal $q(\rvz)$ can be approximated by Monte Carlo:
\begin{equation*}
\begin{split}
    &q(\rvz) =\E_{q(\rvx)}[q(\rvz|\rvx)]\approx \frac{1}{B}\sum_{i=1}^B q(\rvz|\vx^{(i)})\\
    &=\frac{1}{B}\sum_{i=1}^B\frac{1}{(2\pi)^{J/2}\prod_{j=1}^J\sigma_j(\vx^{(i)})}\exp\left\{-\frac{1}{2}\sum_{j=1}^J(\rz_j-\mu_j(\vx^{(i)}))^2/\sigma_j^{2}(\vx^{(i)})\right\}
\end{split}
\end{equation*}
This density is a mixture of Gaussians with $B$ components, and its marginal of $j$-th factor $\rz_j$ is:
\begin{equation*}
\begin{split}
    q(\rz_j)\approx\frac{1}{B}\sum_{i=1}^B\frac{1}{(2\pi)^{1/2}\sigma_j(\vx^{(i)})}\exp\left\{-\frac{1}{2}(\rz_j-\mu_j(\vx^{(i)}))^2/\sigma^2_j(\vx^{(i)})\right\}
\end{split}
\end{equation*}
Hence the log-ratio of $q(\rvz)$ and $\prod_{j=1}^Jq(\rz_j)$ can be approximated as follows:
\begin{equation*}
\begin{split}
    r(\rvz)\equiv\log\frac{q(\rvz)}{\prod_{j=1}^Jq(\rz_j)}
    \approx\frac{\frac{1}{B}\sum_{i=1}^B\frac{1}{(2\pi)^{J/2}\prod_{j=1}^J\sigma_j(\vx^{(i)})}\exp\left\{-\frac{1}{2}\sum_{j=1}^J\frac{(\rz_j-\mu_j(\vx^{(i)}))^2}{\sigma_j^{2}(\vx^{(i)})}\right\}}{\prod_{j=1}^J\frac{1}{B}\sum_{i=1}^B\frac{1}{(2\pi)^{1/2}\sigma_j(\vx^{(i)})}\exp\left\{-\frac{1}{2}\frac{(\rz_j-\mu_j(\vx^{(i)}))^2}{\sigma^2_j(\vx^{(i)})}\right\}}
\end{split}
\end{equation*}
Finally we sample $B'$ representations from each approximate posterior $q(\rvz|\vx^{(i)})$ by reparameterization, and hence totally obtain $BB'$ representations $\{\vz^{(ik)}\}_{i=1,k=1}^{B,B'}$. Eventually, the estimate of total correlation is:
\begin{equation*}
    TC(\rvz)\equiv\KL\left(q(\rvz)\Vert\prod_{j=1}^J q(\rz_j)\right)=\E_{q(\rvz)}[r(\rvz)]\approx\frac{1}{BB'}\sum_{i=1}^B\sum_{k=1}^{B'} r(\vz^{(ik)})
\end{equation*}

The estimate of total correlation above is robust when $B$ and $B'$ is big enough. In experiments, we set $B=64$ and $B'=30$, which is sufficient to produce precise and robust estimate in our considered data sets. As shown in Figure~\ref{fig:Metrics_J=10}, the total correlations are positively correlated to the KL terms in VAEs, which demonstrates the validity of our estimate.

\subsection*{E$\quad$ Disentanglement in InfoGAN}
In this section, we aim at proving that penalizing the mutual information term in InfoGAN with factorized approximate posterior enhances the conditional independence in decoder.

First, we begin with briefly introducing InfoGAN. InfoGAN decomposes the latent variables of decoder (or generator) into three parts $\rvz=(\rvz^d,\rvz^c,\rvz^n)$: $\rvz^d$ is a one hot vector to retain the information of category; $\rvz^c$ is a vector of continuous factors, and each factor is expected to retain the information of one factor variation in data; $\rvz^n$ is noise vector. Then InfoGAN penalizes two mutual information terms: $-I(\rvx;\rvz^d)$ and $-I(\rvx;\rvz^c)$ (maximizes the mutual information). Note that the prior in decoder $p(\rvz)$ is fixed, and hence the penalties are equivalent to $H(\rvz_d|\rvx)$ and $H(\rvz^c|\rvx)$, respectively. To estimate the conditional entropy terms, the posteriors $p(\rvz^d|\rvx)$ and $p(\rvz^c|\rvx)$ should be given. InfoGAN involves an auxiliary network $Q(\cdot)$ to approximate the posteriors by variational approach, in which each data point $\vx$ is encoded into two approximate posteriors $Q(\rvz^d|\vx)$ and $Q(\rvz^c|\vx)$. The approximate posterior for continuous factors $Q(\rvz^c|\rvx)$ is set as a factorized Gaussian, which is vital for conditional independence in decoder as the following discussion shown.

As we introduced above, InfoGAN penalizes $H(\rvz^c|\rvx)$ with a factorized approximate posterior $Q(\rvz^d|\vx)=\prod_{j=1}^JQ(\rz^c_j|\rvx)$. Utilizing this factorized form, the conditional entropy term $H(\rvz^c|\rvx)$ can be decomposed into three terms:
\begin{small}
\begin{equation*}
\begin{split}
    &H(\rvz^c|\rvx)=-\E_{p(\rvx)}\left[\E_{p(\rvz^c|\rvx)}\left[\log p(\rvz^c|\rvx)\right]\right]\\
    =&-\E_{p(\rvx)}\left[\E_{p(\rvz^c|\rvx)}\left[\log p(\rvz^c|\rvx)-\log \prod_{j=1}^Jp(\rz^c_j|\rvx)+\log \prod_{j=1}^Jp(\rz^c_j|\rvx)-\log \prod_{j=1}^JQ(\rz^c_j|\rvx)+\log Q(\rvz^c|\rvx)\right]\right]\\
    =&-\E_{p(\rvx)}\left[\E_{p(\rvz^c|\rvx)}\left[\log\frac{p(\rvz^c|\rvx)}{\prod_{j=1}^Jp(\rz^c_j|\rvx)}\right]+\E_{p(\rvz^c|\rvx)}\left[\log\frac{\prod_{j=1}^Jp(\rz^c_j|\rvx)}{\prod_{j=1}^JQ(\rz^c_j|\rvx)}\right]+\E_{p(\rvz^c|\rvx)}\left[\log Q(\rvz^c|\rvx)\right]\right]\\
    =&-\E_{p(\rvx)}\left[\KL\left(p(\rvz^c|\rvx)\Vert\prod_{j=1}^Jp(\rz^c_j|\rvx)\right)+\sum_{j=1}^J\KL(p(\rz^c_j|\rvx)\Vert Q(\rz^c_j|\rvx))\right]-\E_{p(\rvz^c,\rvx)}\left[\log Q(\rvz^c|\rvx)\right]
\end{split}
\end{equation*}
\end{small}
The third term above $-\E_{p(\rvz^c,\rvx)}\left[\log Q(\rvz^c|\rvx)\right]$ is exactly the true penalty in InfoGAN. Note that the training target of $Q(\cdot)$ net is to minimize $-\E_{p(\rvz^c,\rvx)}\left[\log Q(\rvz^c|\rvx)\right]$, which is also one of the training targets of decoder, so this term will well approximate $H(\rvz^c|\rvx)$ when the $Q$ net converges. In this case, the first two terms should be nearly zero, which leads to conditional independence in decoder. Therefore, factorized approximate posterior in InfoGAN is an inductive bias for conditional independence in decoder.

\textbf{Additional experimental results:} To demonstrate that the correlated approximate posterior has no obvious impact on the convergences of InfoGAN, here we show the loss curves in the training process of InfoGAN.
\begin{figure}[htb!]
%\vspace{-20pt}
\setlength{\belowcaptionskip}{-0.7cm}
\centering
    \subfigure[$\sigma=0$]{
	\includegraphics[width=4.5cm,height=3.5cm]{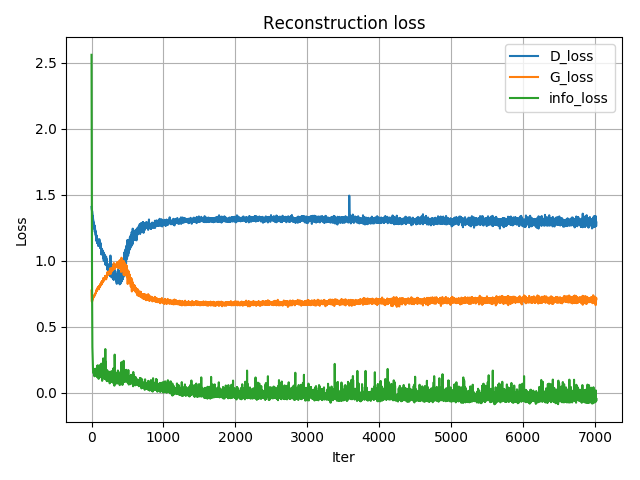}}
	\subfigure[$\sigma=0.9$]{
	\includegraphics[width=4.5cm,height=3.5cm]{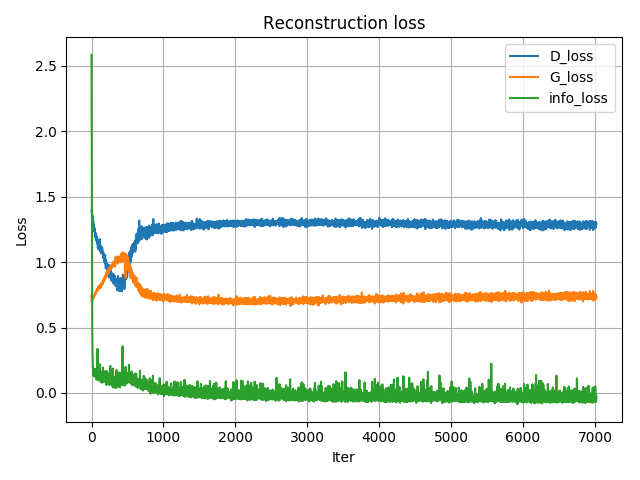}}
	\vspace{-10pt}
	\caption{\textbf{Loss curves in the training process of InfoGAN.} The blue curve, original curve and green curve are the losses of discriminator, decoder (generator) and the mutual information penalty, respectively. Models with $\sigma=0$ and $\sigma=0.9$ both converge well.}
\end{figure}

\subsection*{F$\quad$ Experimental Results in VAEs with $20$ Factors}
To demonstrate that the decline of disentanglement by involving approximate posterior is stable when the number of factors $J$ varies, we perform experiments on VAEs with $J=20$, and show the MIG scores here.
\begin{figure}[tb!]
%\vspace{-20pt}
%\setlength{\belowcaptionskip}{-0.0cm}
\centering
    \subfigure[dSprites]{
	\includegraphics[width=4.5cm,height=3.5cm]{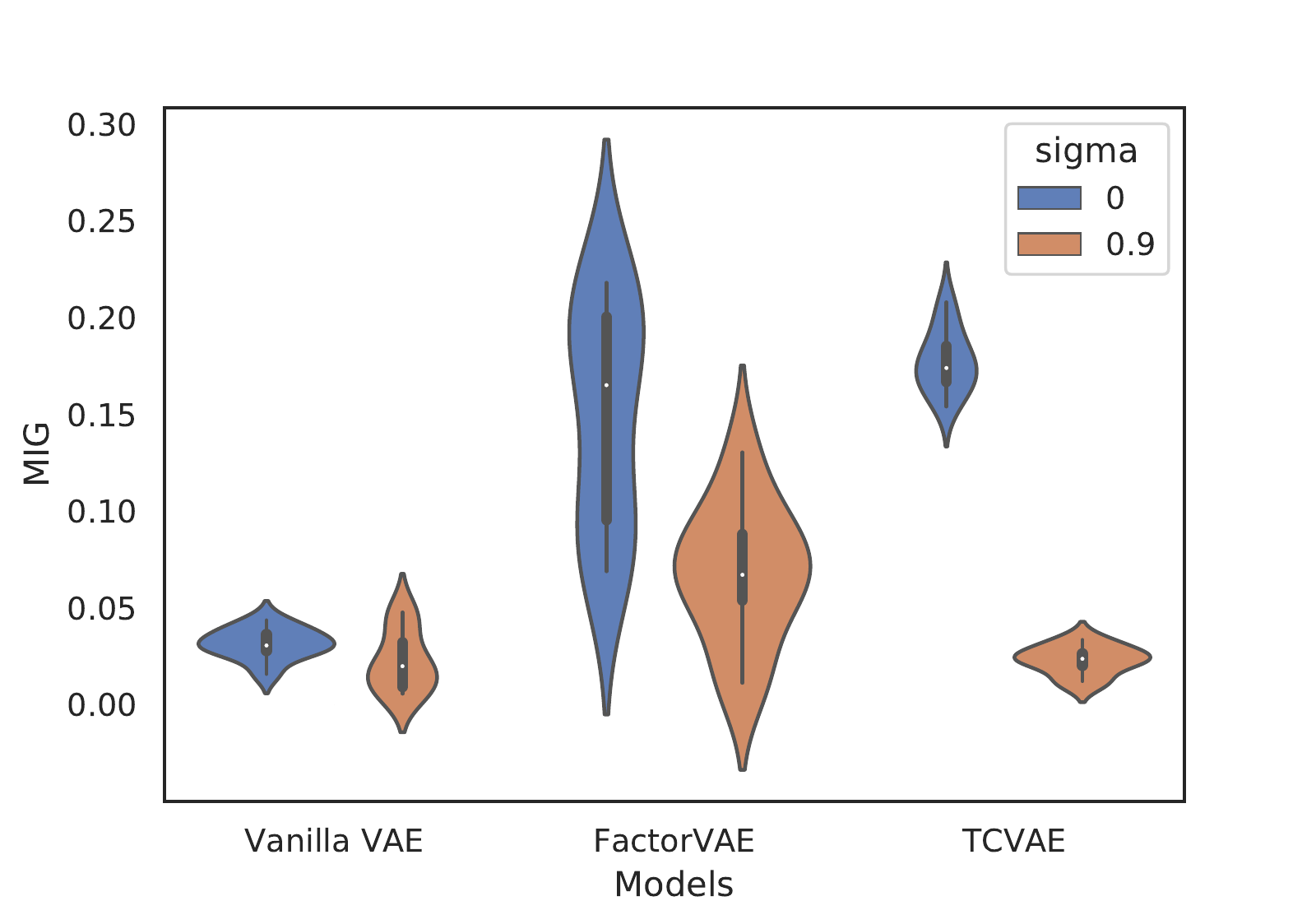}}
	\subfigure[SmallNORB]{
	\includegraphics[width=4.5cm,height=3.5cm]{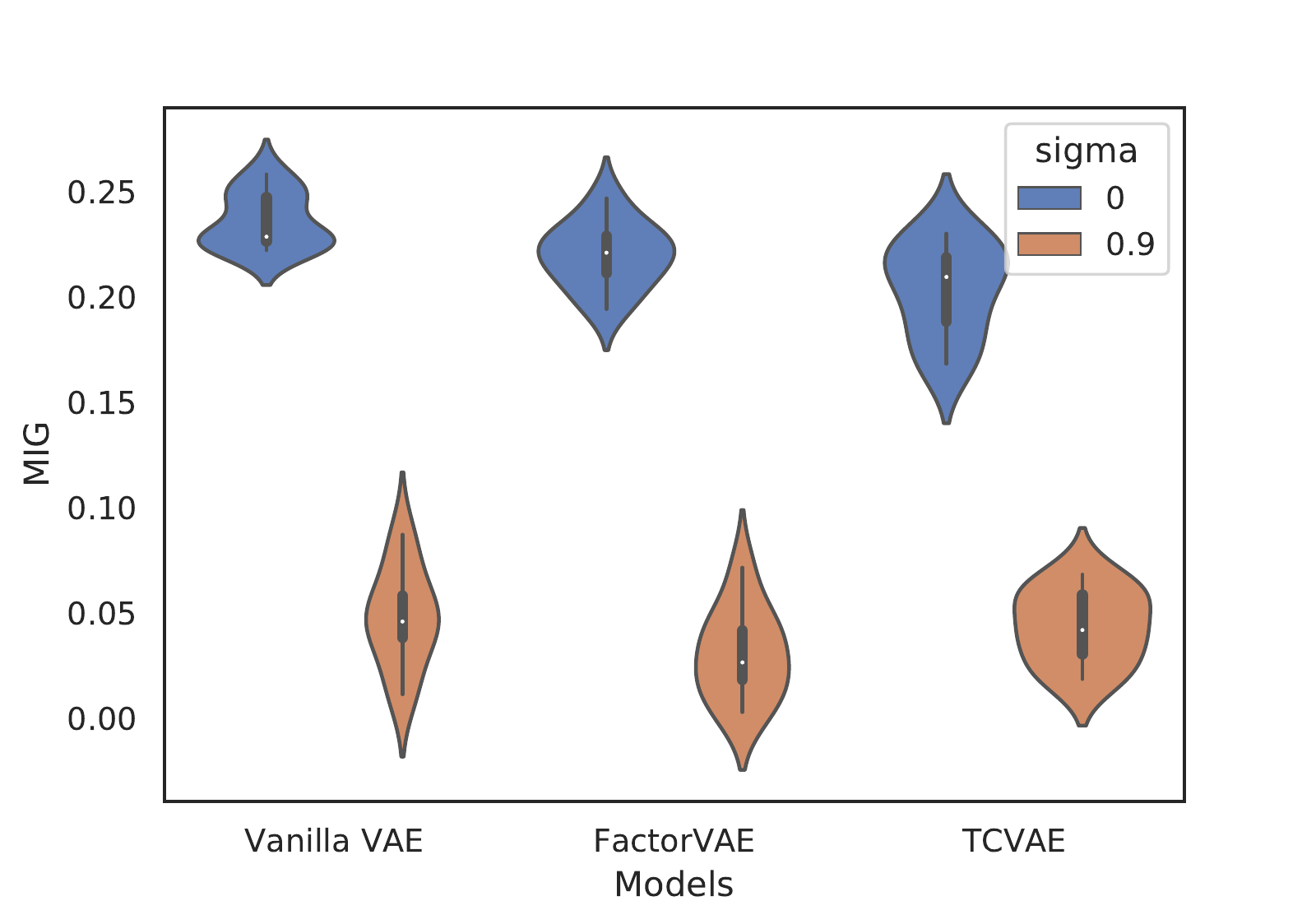}}
	\subfigure[Cars3D]{
	\includegraphics[width=4.5cm,height=3.5cm]{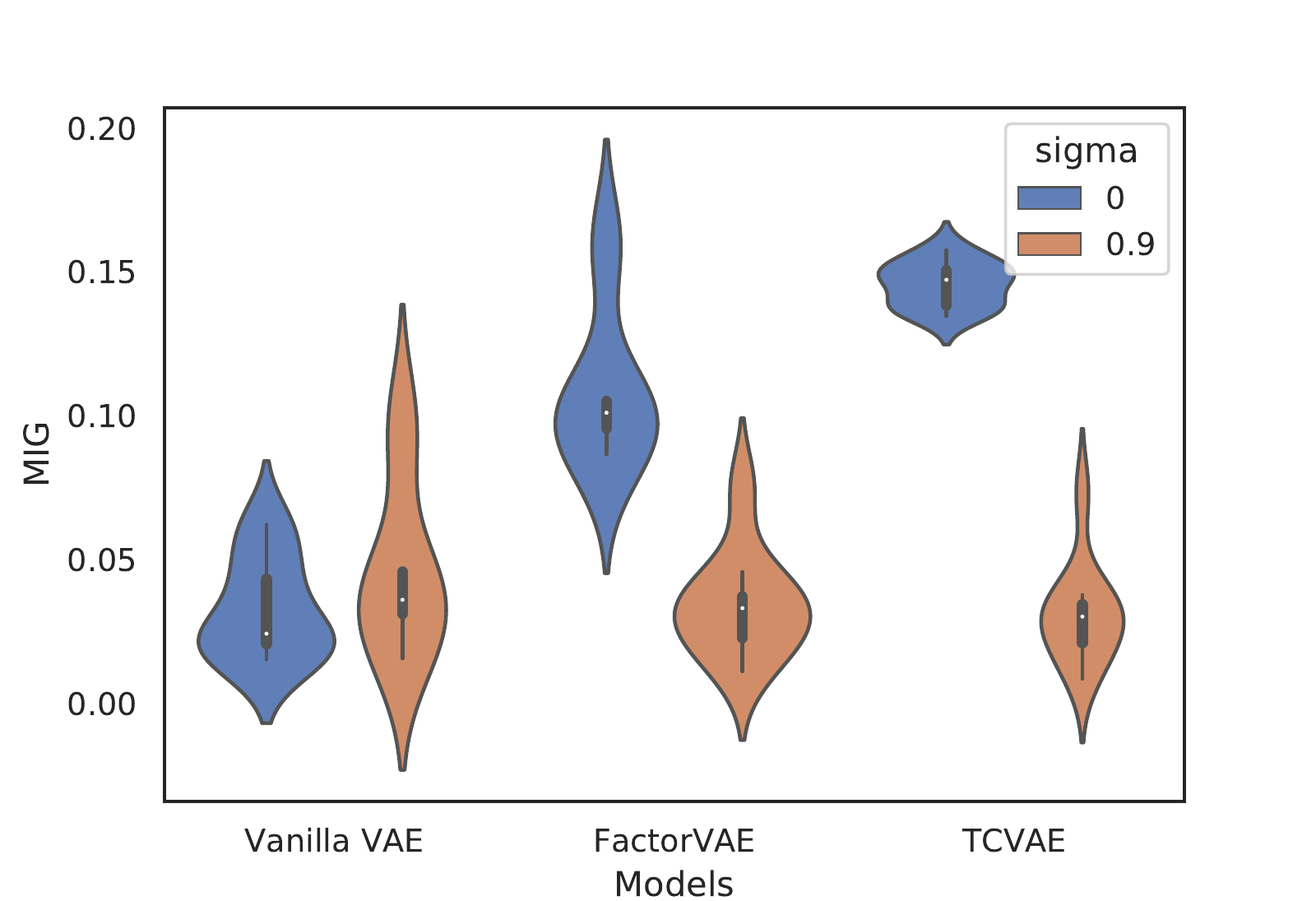}}
	\vspace{-10pt}
	\caption{\textbf{MIG of vanilla VAE, FactorVAE and TCVAE on dSprites, SmallNORR and Cars3D when $J=20$.} The blue points are results with $\sigma=0$, while the orange points are for $\sigma=0.9$.}
\end{figure}
When $J=20$, the decline of disentanglement measured by MIG when $\sigma$ varies from $0$ to $0.9$ is still consistent across models and data sets, and even more obvious than the case of $J=10$ in FactorVAE and TCVAE compared with Figure~\ref{fig:MIG_J=10}. The anomaly of vanilla VAE on Cars3D is slight, which might be due to the low MIG score of vanilla VAE and the statistical fluctuations, as well as the weakness of independence of factors in vanilla VAE.

\bibliographystyle{plainnat}
\end{document}